\documentclass[11pt]{article}

\usepackage[final]{acl}
\usepackage{times}
\usepackage{latexsym}
\usepackage[T1]{fontenc}
\usepackage{amsmath,amssymb}
\usepackage{booktabs}
\usepackage{graphicx}
\graphicspath{{figures/}}
\usepackage{float}
\usepackage{microtype}
\usepackage{pifont}

\hypersetup{colorlinks=true,
  linkcolor=[RGB]{120,20,20},
  citecolor=[RGB]{0,68,120},
  urlcolor=[RGB]{0,68,120},
  pdftitle={Fusion Embedding: A Unified Embedding Space for Text, Image,
    Video, and Audio},
  pdfauthor={Abdul Basit Tonmoy, Kazi Fardinul Hoque, Md. Shahrier Islam Arham, Arman Luthra}}

\newcommand{\family}{Fusion Embedding}
\newcommand{\fea}{\texttt{fusion-embedding-1}}
\newcommand{\feb}{\texttt{fusion-embedding-2}}
\newcommand{\feaURL}{https://huggingface.co/EximiusLabs/fusion-embedding-1-2b-preview}
\newcommand{\febURL}{https://huggingface.co/EximiusLabs/fusion-embedding-2-2b-preview}
\newcommand{\ratk}[1]{R@#1}

\title{\family: A Unified Embedding Space for Text, Image, Video, and Audio}

\author{%
  Abdul Basit Tonmoy$^{1,2,3}$\thanks{\,Corresponding author.} \quad
  Kazi Fardinul Hoque$^{2}$ \quad
  Md.\ Shahrier Islam Arham$^{1,2}$ \quad
  Arman Luthra$^{1,2}$ \\[3pt]
  $^{1}$Eximius Labs \quad $^{2}$Wabash College \quad
  $^{3}$Skop Intelligence Co. \\[3pt]
  \texttt{atonmoy27@wabash.edu} \quad \texttt{kfhfar@amazon.com} \\[2pt]
  \texttt{marham27@wabash.edu} \quad \texttt{aluthra26@wabash.edu}}

\date{}

\begin{document}
\maketitle

\begin{abstract}
A single embedding space that covers text, images, video, and audio lets one
index serve every query a user can pose. Embedding models built on
vision--language backbones now lead text/image/video retrieval benchmarks
but lack audio entirely, while audio--text retrieval is led by specialist
systems that serve no other modality. We present the \family{} family, which
adds audio to a frozen vision--language embedding base whose parameters are
never updated: generation~1 (\fea) trains only a 16.4M-parameter connector
between a frozen audio tower and the frozen base, and generation~2 (\feb)
adds modality-gated deep adapters (44.2M parameters) whose branch
never executes on text, image, or video inputs: their outputs are
bit-for-bit those of the released base, verified after every training run.
Because the base already binds text, images, and video, aligning audio
to text alone makes audio--image retrieval \emph{emerge}, with zero paired
audio--visual training data. Alongside the recipe we map its design space
with controlled negative results (rewriting training captions with an
LLM, substituting a leaderboard-stronger audio tower, and widening the
connector each \emph{reduce} retrieval) and with training-protocol
findings that we expect to transfer to any frozen decoder-LM embedding
backbone. Both generations train in hours on a single GPU. Weights, code,
and the evaluation harness
are openly released.
\end{abstract}

\section{Introduction}
\label{sec:intro}

Retrieval systems increasingly need a \emph{single} vector space covering
every modality a user might store or query: text, images, video, and audio.
Contrastive dual encoders trained per modality pair, such as CLIP~\citep{radford2021clip}
for vision--language and the CLAP family for
audio--language~\citep{wu2023laionclap,mei2023wavcaps,zhu2024cacophony,niizumi2025m2dclap},
achieve strong in-pair retrieval but each model spans only two
modalities. Unified-space models such as ImageBind~\citep{girdhar2023imagebind}
bind many modalities to a pivot modality, demonstrating that alignment
between two modalities can emerge through a shared third. Meanwhile,
embedding models built on large vision--language backbones now lead
text/image/video retrieval benchmarks such as
MMEB-V2~\citep{jiang2024vlm2vec,meng2025vlm2vecv2}, but lack audio entirely.

Two routes to adding the missing modality are emerging, and they trade
different things. \emph{Audio-native specialist systems} adapt a multimodal
LLM from the inside: OEA~\citep{omniembedaudio2026} trains LoRA adapters in
the backbone attention of an audio-native LLM, and
AuroLA~\citep{aurola2026} trains 418M LoRA parameters inside both the audio
tower and the LLM of Qwen2.5-Omni-7B, adding a second 7B model as a
re-ranker. These systems set the pace on audio--text benchmarks, and they
serve exactly one modality pair, with backbones that are no longer the
models their ecosystems shipped. \emph{Frozen-base fusion}, the route taken
here, holds the deployed model fixed and attaches the new modality from the
outside. The trade is capability for invariance: every embedding a
deployment has already computed and indexed with the base model remains
exactly valid, and the base's published text/image/video benchmark results
carry over unchanged; not approximately, but by construction.

This report describes the two generations of the \family{} family,
\href{\feaURL}{\fea} and \href{\febURL}{\feb}, both built on a
byte-frozen Qwen3-VL-Embedding-2B~\citep{qwen3vlembedding} base and the
frozen audio tower of Qwen2.5-Omni~\citep{xu2025qwen25omni}.
Generation~1 establishes the space: a 16.4M-parameter perceiver-resampler
connector, trained contrastively against cached frozen-text targets, is
enough to make audio a first-class modality: AudioCaps audio-to-text
\ratk{10} 0.741 with both towers frozen, Clotho transfer strictly zero-shot,
and audio-to-image retrieval at 29$\times$ chance with no audio--visual
supervision. Generation~2 asks where the remaining headroom lives. Three
controlled negative results (\S\ref{sec:negative}) locate it in the frozen
LM's ability to \emph{process} audio tokens, not in the data or the
connector; \feb{} therefore adds bottleneck adapters at every decoder layer,
hard-gated to audio inputs so that non-audio computation is bitwise
untouched. The adapters improve every text-to-audio benchmark (AudioCaps
\ratk{10} 0.775) while the invariance guarantee holds exactly.

Our contributions:
\begin{itemize}
  \item \emph{System.} A two-generation recipe for extending a frozen
  vision--language embedding base with audio: a 16.4M connector
  (generation~1), then modality-gated deep adapters (+44.2M, generation~2)
  whose gate guarantees that non-audio inputs execute the base's
  computation unmodified (\S\ref{sec:guarantee}). Connector-only systems
  preserve the base structurally; the gate extends the same exactness to
  \emph{in-layer} capacity, where prior methods only mitigate drift:
  unmerged LoRA preserves base \emph{parameters} but executes on the text
  path, and zero-initialized expansion trains away from identity.
  \item \emph{Emergent Cross-Modal Retrieval.} With zero audio--image
  training pairs, audio-to-image retrieval on VGGSound-696 reaches \ratk{10}
  0.418 (29$\times$ chance; 0.443 for the strongest generation-2
  checkpoint, under a different readout template; see Table~\ref{tab:fe2}). Head-to-head under one protocol, each of ImageBind and
  \fea{} wins its supervised pair and \fea{} wins the two remaining pairs;
  \fea{} beats LanguageBind on all three pairs including LanguageBind's
  supervised one, and we trace that failure to measurable divergence of its
  branch text towers (\S\ref{sec:crossmodal}).
  \item \emph{Negative results.} Controlled experiments in which
  LLM-rewritten training captions \emph{reduce} retrieval despite being
  cleaner; a sound-event audio tower that dominates shallow-projection
  comparisons loses by 16 \ratk{10} points inside a frozen-LLM splice; and
  wider connectors lose by memorizing. Together with a +14.5-point
  training-protocol effect and a loss-floor audit, these map the design
  space of frozen-backbone audio fusion (\S\ref{sec:negative}).
  \item \emph{Release.} Open weights for both generations with pinned
  revision tags, Apache-2.0 training and evaluation code, and result
  records for every reported number (\S\ref{sec:conclusion}).
\end{itemize}

\section{Related Work}
\label{sec:related}

\paragraph{Contrastive audio--text pretraining (CLAP family).}
LAION-CLAP~\citep{wu2023laionclap}, WavCaps~\citep{mei2023wavcaps},
Cacophony~\citep{zhu2024cacophony}, and M2D-CLAP~\citep{niizumi2025m2dclap}
train audio and text encoders end-to-end on hundreds of thousands of
audio--caption pairs and set the AudioCaps retrieval state of the art among
dual encoders (audio-to-text \ratk{10} 0.906--0.928). These models fine-tune
both towers and include in-domain AudioCaps training; M2D-CLAP additionally
includes Clotho in its training data, so its Clotho rows are not zero-shot.
WavCaps also demonstrated the value of large LLM-assisted caption corpora,
and AudioSetCaps~\citep{bai2024audiosetcaps} scaled this recipe to 1.9M
pairs; notably, its authors report that in-domain AudioCaps fine-tuning
after pretraining is worth +13--15 \ratk{1} points, and that a 478K
high-quality-caption subset outperforms a similarly sized mix of
WavCaps+AudioCaps+Clotho; caption quality wins at equal scale.
GLAP~\citep{dinkel2025glap} compares five frozen audio encoders under one
shared contrastive recipe and selects Dasheng~\citep{dinkel2024dasheng} as
the most versatile, with Whisper-family encoders trailing sound-event
encoders by $\sim$9--12 mAP@10 on sound retrieval; our
tower-swap experiment shows this ranking does \emph{not} transfer to frozen-LLM splicing
(\S\ref{sec:negative}). On the data side,
AudioCaps~2.0~\citep{audiocaps2_2025} (91{,}256 train / 4{,}875 test pairs,
audio distribution gated behind a request process) doubles the in-domain
pool; we verified its train split is disjoint from the v1 test protocol we
evaluate on (0 of our 883 evaluation clips appear in its train CSV), making
it a natural extension of our fine-tuning stage. Our models differ from
this entire family in that \emph{neither} tower is trained.

\paragraph{Audio-native specialist systems.}
Recent work builds audio--text retrieval on multimodal LLM backbones.
OEA~\citep{omniembedaudio2026} is the closest system to generation~1 in
trainable-parameter budget: it keeps an audio-native multimodal LLM
backbone frozen and trains LoRA adapters on the attention layers plus
projection heads (11--16M trainable parameters against our 16.4M
connector), reaching AudioCaps text-to-audio \ratk{1} 0.389. The
architectural difference is decisive for our setting: OEA's LoRA sits
\emph{inside} the backbone attention, so the text pathway is adapted too
and the shared backbone is no longer byte-identical to its release,
which is exactly the property our design exists to preserve. Two further
observations from OEA inform our analysis: its 7B backbone performs on par
with its 3B on the audio benchmarks, and its Clotho-trained variants are not
zero-shot on Clotho. AuroLA~\citep{aurola2026} trains LoRA in both towers
of Qwen2.5-Omni-7B (418M trainable parameters) on a curated 1.4M-pair
corpus and adds a second 7B model for pairwise re-ranking, reaching
AudioCaps \ratk{1} 0.656 (A$\to$T) / 0.510 (T$\to$A) as a two-pass system;
its scaling ablation indicates that large gains lie exactly in the corpus
range above ours. Its Hybrid-NCE loss (soft positive sets with weighted
negatives) is convergent evidence for the soft-label and
false-negative-masking terms we adopt at large corpus scale
(\S\ref{sec:loss}). Both lines validate LLM backbones for audio retrieval
while adapting far more than the audio path, and neither serves, nor
evaluates, any modality beyond audio and text.

\paragraph{Unified embedding spaces.}
ImageBind~\citep{girdhar2023imagebind} binds six modalities to images and
established that cross-modal alignment emerges between modalities never
paired in training. We use the same emergence mechanism with text as the
pivot, but in the opposite regime: instead of training modality towers to
bind to a pivot encoder, we freeze an existing multi-modality base and
train only the new modality's path, which preserves the base's benchmark
performance exactly. LanguageBind~\citep{zhu2024languagebind} binds
modality branches to language by fine-tuning per-branch copies of a text
tower, a design whose consequences we measure in
\S\ref{sec:crossmodal}. ONE-PEACE~\citep{wang2023onepeace} trains a 4B
three-modality (vision/audio/language) model from scratch and remains the
strongest unified space on audio--text retrieval, with every parameter
trained and its retrieval rows fine-tuned on the merged
AudioCaps+Clotho+MACS train splits (Table~\ref{tab:audiocaps}). A separate
\emph{space-stitching} line (OmniBind, \citealp{wang2025omnibind};
FreeBind, \citealp{wang2024freebind}) binds many pre-trained specialist
spaces with small learned projectors and routers; these systems report
AudioCaps/Clotho \ratk{1}/\ratk{5} only (retrieval direction unstated), and
the bound experts include checkpoints trained on AudioCaps/Clotho train
data, so their ``zero-shot'' applies to the stitching stage rather than
the underlying spaces. Matryoshka representation
learning~\citep{kusupati2022mrl} provides the nested-truncation property we
preserve end-to-end; our contrastive objective is applied at every rung of
the base's ladder, and our whitening is deliberately diagonal so truncation
and whitening commute.

\paragraph{Positioning.}
Concurrent with jina-embeddings-v5-omni~\citep{honicke2026jinaomni}, which
composes a frozen \emph{text} embedding backbone with frozen vision and audio
encoders and trains only connecting components, \fea{} is, to our knowledge,
the first to extend an already \emph{vision--language} embedding model
(Qwen3-VL-Embedding-2B) to audio with both the base and the audio tower frozen
and only a connector trained: a fourth modality added to an existing retriever
rather than a multimodal space built from a text base. \feb{} adds audio
capacity \emph{inside} the frozen backbone's layers, gated by modality. Unlike
gated cross-attention~\citep{alayrac2022flamingo} or block
expansion~\citep{wu2024llamapro}, which preserve the original behavior only
approximately or only at initialization, the gate makes the text, image, and
video outputs \emph{bitwise-identical} to the frozen base---a guarantee we
verify on the released checkpoint, and one we are not aware of any prior
multimodal embedding model providing across several preserved modalities at
once. On AudioCaps audio-to-text \ratk{10}, \fea{} and \feb{} (${\sim}0.74$)
are the frontier of the both-towers-frozen class; the next such system is
jina-embeddings-v5-omni at 0.672, and every model above them trains at least
one tower. ONE-PEACE reaches 0.92, but in a different class: all 4B
parameters trained, with in-domain fine-tuning. The audio-native
specialist systems above sit well higher on in-domain audio--text \ratk{1}
(\S\ref{sec:main-results}); they are single-pair systems, and the
comparison across system classes is drawn explicitly in
Table~\ref{tab:audiocaps}.

\section{The Model Family}
\label{sec:method}

\begin{figure*}[t]
\centering
\includegraphics[width=\textwidth]{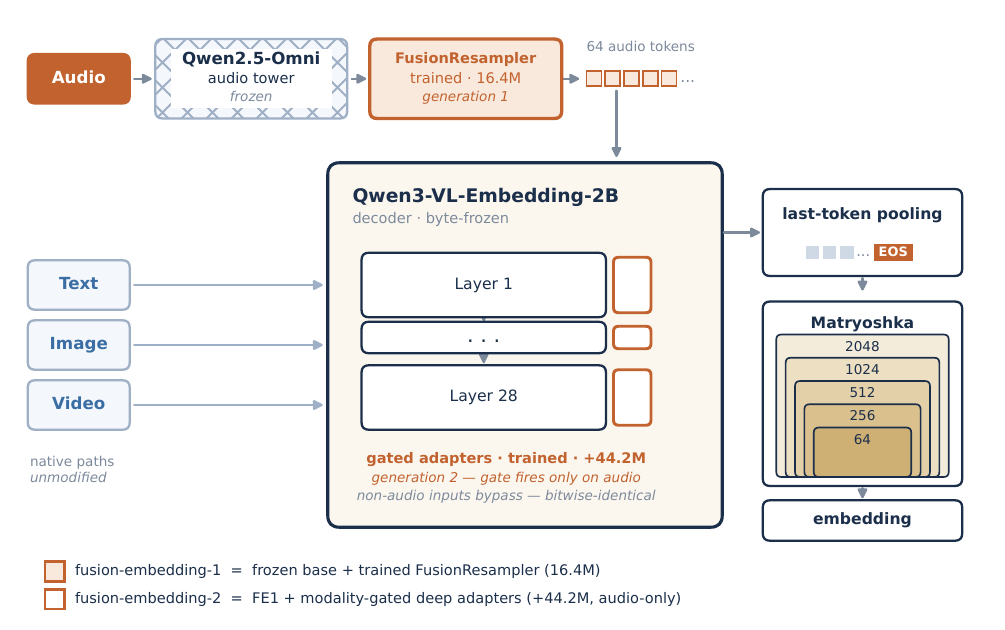}
\caption{The \family{} family. Both generations share the frozen stack:
a byte-frozen Qwen3-VL-Embedding-2B base (last-token pooling, Matryoshka
output ladder) and the frozen Qwen2.5-Omni audio tower. Generation~1
(\fea) trains only the FusionResampler (16.4M parameters, width 384, 64
latent queries), which maps audio-tower frames into the base's input
embedding space at placeholder token positions. Generation~2 (\feb) keeps
that connector and adds a bottleneck adapter to the residual stream of each
of the 28 frozen decoder layers (44.2M parameters), gated so the adapter
branch executes only while audio is being encoded: text, image, and video
inputs take the unmodified frozen path and their outputs are bit-for-bit
identical to the released base. Training updates only the orange
components; every run asserts \texttt{base\_drift} $= 0$ afterwards.}
\label{fig:architecture}
\end{figure*}

Figure~\ref{fig:architecture} shows the family: two frozen towers, a
trained connector (generation~1), gated per-layer adapters (generation~2),
and the base's own read-out. Table~\ref{tab:spec} summarizes the released
models.

\begin{table}[t]
\centering
\small
\begin{tabular}{lcc}
\toprule
 & Gen.~1 & Gen.~2 \\
\midrule
Frozen base & \multicolumn{2}{c}{Qwen3-VL-Emb.-2B} \\
Frozen audio tower & \multicolumn{2}{c}{Qwen2.5-Omni} \\
Trained parameters & 16.4M & 60.6M \\
Base params.\ updated & 0 & 0 \\
Modalities (one space) & \multicolumn{2}{c}{text, image, video, audio} \\
Embedding dim.\ (MRL) & \multicolumn{2}{c}{2048 $\to$ 64 nested} \\
Instruction conditioning & \ding{51} & \ding{51} \\
Bitwise base invariance & \ding{51} & \ding{51} \\
In-layer audio capacity & \ding{55} & \ding{51} \\
\bottomrule
\end{tabular}
\caption{The released family at a glance (Gen.~1 $=$ \fea, Gen.~2 $=$
\feb). ``Bitwise base invariance'': inputs containing no audio execute the
unmodified base computation, so their outputs are bit-for-bit those of the
released base model under matched execution (\S\ref{sec:guarantee}). Both
models keep the base's Matryoshka ladder and instruction interface.}
\label{tab:spec}
\end{table}

\subsection{Generation 1: a connector and nothing else}
\label{sec:arch}

\paragraph{Frozen components.}
The base is Qwen3-VL-Embedding-2B~\citep{qwen3vlembedding}: a decoder-LM
embedding model with hidden width $d_{\mathrm{llm}}=2048$, last-token (EOS)
pooling, instruction conditioning via a chat template, and a Matryoshka
ladder $\{2048, 1536, 1024, 512, 256, 128, 64\}$. The audio tower is the
audio encoder of Qwen2.5-Omni-7B~\citep{xu2025qwen25omni}, a
Whisper-large-v3-derived~\citep{radford2022whisper} encoder consuming
128-bin log-mel spectrograms of 30\,s windows at 16\,kHz. We tap the
tower's \emph{post-projection} output: packed frames of dimension 3584 at
$\sim$25 frames per second; an early A/B confirmed this tap outperforms the
1280-d pre-projection encoder states. Both components are frozen: they are
set to evaluation mode, gradients are disabled, and a regression guard
(\S\ref{sec:guarantee}) verifies at the end of every run that no base
parameter changed.

\paragraph{FusionResampler (the only trained module).}
The connector is a Flamingo-style perceiver
resampler~\citep{alayrac2022flamingo} operating at an internal bottleneck
width $d_r = 384$: an input projection $3584 \to 384$ with sinusoidal
temporal positions, $N = 64$ learnable latent queries processed by $L = 6$
pre-norm blocks (latent self-attention $\to$ cross-attention over the audio
frames with key-padding masks $\to$ 4$\times$ FFN), and an output
projection $384 \to 2048$ followed by LayerNorm. This totals 16.4M trained
parameters, under 1\% of the 2B base. A learnable temperature (initialized
to $\log(1/0.07)$, clamped at $\log 100$) is the only other trained scalar.

\paragraph{Injection at placeholder positions.}
The connector's $N$ output tokens overwrite the input embeddings of $N$
placeholder tokens in the base's input stream, an inert special token of
the base vocabulary reserved for this purpose, exactly mirroring the
base's own image-token mechanism. The frozen LLM then runs its ordinary
forward pass over the spliced sequence, and the embedding is read out by
the base's own last-token pooling, truncated to a Matryoshka rung, and
L2-normalized. Audio thus conforms to the base's read-out conventions in
every respect; nothing about the base's text, image, or video paths is
altered.

\subsection{Generation 2: modality-gated deep adapters}
\label{sec:adapters-arch}

Generation~1 routes all audio understanding through a 16.4M input-side
connector into layers trained only on text and images. The negative
results of \S\ref{sec:negative} locate the remaining bottleneck in exactly
those layers; OEA reaches near-specialist retrieval at a comparable
trainable budget precisely because its LoRA adapts the layers themselves
(a route closed to us, since LoRA touches every token and forfeits
backbone byte-identity).

\feb{} adds the in-layer capacity without giving up the guarantee. At each
of the base's 28 decoder layers, a bottleneck adapter
($\mathrm{LayerNorm} \to 2048{\times}384 \to \mathrm{SiLU} \to
384{\times}2048$; 44.2M parameters total) is attached to the residual
stream, \emph{hard-gated to audio encoding}: for any text, image, or video
input the hook returns the frozen layer's output before any adapter
arithmetic executes, so non-audio forwards are bitwise identical to the
unmodified base. The adapter up-projections are zero-initialized, making a
fresh adapter stack the exact identity, so training starts at the
generation-1 architecture. One implementation constraint matters in
practice: with gradient checkpointing, layer forwards re-run during the
backward pass, so the gate must be held open across forward \emph{and}
backward of each audio step (non-reentrant checkpointing detects a mismatch
loudly); we additionally hard-fail any text encode issued under an open
gate, so the frozen text targets cannot be silently corrupted.

\subsection{The invariance guarantee}
\label{sec:guarantee}

The family's defining property is that the base model is provably
unmodified, at two levels. \emph{Parameters}: a regression guard snapshots
every base parameter before training and asserts bitwise equality after; a
run with nonzero drift fails, and the released checkpoints contain only the
connector, adapters, a temperature, and normalization statistics.
\emph{Computation}: in generation~2, the adapter branch does not execute
outside the audio gate, so text, image, and video forwards are the frozen
base's forwards: the same operations on the same weights. Stated
precisely, the property is \emph{computation-graph identity for inputs
containing no audio tokens}: a mixed sequence that does contain audio is
adapted as a whole, including its text tokens, while any input without
audio never touches an adapter. Bitwise output equality is the verified
corollary, checked by unit test and release smoke test under matched
kernels, precision, and execution (\S\ref{sec:invariance}); we do not
claim bit equality across arbitrary hardware, where floating-point
kernels themselves differ. This converts ``the base's MMEB-V2
text/image/video results are preserved'' from an empirical claim requiring
re-benchmarking into a mechanical consequence of the architecture, checked
on every run. It also has a deployment reading: adding audio to an index
built on the base model re-embeds \emph{nothing}; every stored vector
remains exactly valid.

\section{Training Recipe}
\label{sec:recipe}

\subsection{Objective and negatives}
\label{sec:loss}

Let $a_i, t_i \in \mathbb{R}^{2048}$ be the pooled audio and text vectors
of a batch of $B$ pairs. The loss is symmetric
InfoNCE~\citep{oord2018infonce} computed at \emph{every} Matryoshka rung
$D$ (prefix-truncate to $D$, renormalize), weighted equally and summed,
plus a light CORAL covariance penalty~\citep{sun2016coral}
($\lambda = 0.05$) that discourages audio from forming its own cluster:
\begin{equation}
\begin{split}
\mathcal{L} = \sum_{D \in \mathrm{MRL}} w_D \,
\tfrac{1}{2}\bigl[&\mathrm{InfoNCE}_D(a \to t) \\[-4pt]
 &+ \mathrm{InfoNCE}_D(t \to a)\bigr] \\
 + \lambda \,\bigl\| \mathrm{Cov}(a) &- \mathrm{Cov}(t) \bigr\|_F^2 / d^2 .
\end{split}
\end{equation}

\paragraph{Full-corpus frozen-text negative bank.}
The audio-to-text direction augments the in-batch negatives with the
\emph{entire corpus} of cached caption embeddings as shared negatives:
484{,}372 rows in the generation-1 headline run. The classic memory-bank
failure mode is staleness: banked representations drift as their encoder
trains. Here the text tower is frozen, so cached text embeddings are exact
forever; the bank has zero staleness \emph{by construction}. Each
anchor's own caption and its exact-string duplicates are masked out of the
denominator. Text anchors use in-batch audio negatives only (audio
embeddings do change during training and are never banked). Because the
frozen text side is precomputed once into a cache, the bank costs one
matrix multiply per step; the cache itself gave +27\% steps/min and
$-$45\% peak VRAM (43.8\,GB $\to$ 23.9\,GB at batch 128), which is what
makes effective batch 1024 practical on a single accelerator.

\paragraph{Soft labels and false-negative masking at large scale.}
Past $\sim$500K corpus rows the bank itself becomes the constraint
(\S\ref{sec:crowding}): thousands of moderately similar captions crowd the
denominator. Two terms counteract this and are enabled for all runs on
500K+ corpora: a soft-label relaxation of the InfoNCE targets
($\beta = 0.3$) and masking of bank entries whose caption similarity to the
anchor exceeds $\tau = 0.98$ (the threshold at which captions stop being
paraphrases; \S\ref{sec:floor}). In a matched A/B at the 592K scale the two
terms recover +2.6 \ratk{10} over the unmodified objective (0.501 $\to$
0.527 on the in-run probe protocol\footnote{The \emph{in-run probe
protocol} is the automated AudioCaps rescore that every training job runs
at its own precision. Probe numbers are comparable between matched arms
within this paper; release numbers come from the protocol of
\S\ref{sec:protocols} at matched bf16 lineage.}). AuroLA's Hybrid-NCE (soft positive
sets with weighted negatives, of which InfoNCE is a special case)
arrives at the same structure independently~\citep{aurola2026}, which we
read as convergent evidence that plain InfoNCE is the wrong objective at
this corpus scale.

\subsection{The native-protocol effect: +14.5 points}
\label{sec:native}

Decoder-LM embedding models are trained and released with a specific input
protocol. For Qwen3-VL-Embedding this is the chat template with the task
instruction in the system turn, the content in the user turn, and
generation priming, with the embedding pooled at the final position. Our
early runs fed \emph{bare} sequences
(``\{instruction\} \{caption\}<|im\_end|>''), off the base's training
manifold. Rebuilding the caption-embedding cache in the native format and
retraining the connector at otherwise identical configuration improved
AudioCaps audio-to-text \ratk{10} from 0.481 to \textbf{0.626} (+14.5
points; \ratk{1} +73\%) at half the schedule of the previous best. This is the
single largest improvement in this work, exceeding the gains from whitening
(+11.5 in the bare format) and from a 19K$\to$131K data scale-up (+9.3).

\begin{figure}[t]
\centering
\includegraphics[width=\columnwidth]{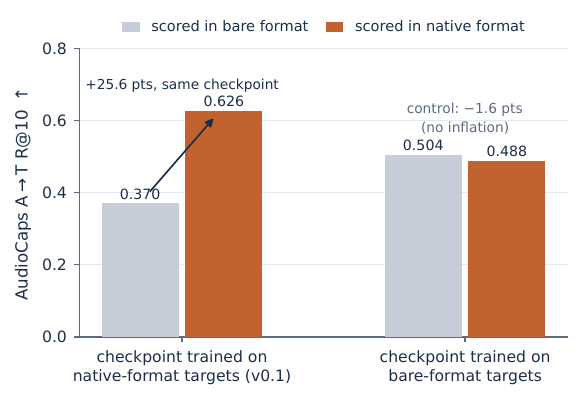}
\caption{Protocol sensitivity (AudioCaps A$\to$T \ratk{10}). The
\emph{same} generation-1 checkpoint reads 25.6 points apart under the bare
vs.\ native input format (left), while the bare-trained control (right)
shows the native format inflates nothing. The deployment lesson: evaluate
a frozen backbone in its training protocol, or silently forfeit
double-digit recall. Both formats are printed verbatim in
Appendix~\ref{app:templates}.}
\label{fig:protocol}
\end{figure}

The gain is not an evaluation artifact, and the sensitivity cuts both ways
(Figure~\ref{fig:protocol}). The v0.1 checkpoint scored under the bare
input format reads AudioCaps \ratk{10} \textbf{0.370}; the same checkpoint
under the native chat-template format reads \textbf{0.626}, a 25.6-point
swing from formatting alone, because a connector trained against
native-format targets is evaluated off its manifold otherwise. The control
rules out score inflation: a checkpoint trained on \emph{bare}-format
targets scores 0.488 under the native format vs.\ 0.504 under its own:
native formatting does not flatter models not trained for it. The general
lesson: \emph{a frozen embedding backbone must be trained against, and
evaluated in, its native input protocol}; off-protocol targets silently
forfeit double-digit recall, and published comparisons between
LLM-backbone embedding models should be read with the input protocol
attached.

\subsection{Diagonal, Matryoshka-safe text whitening}
\label{sec:whitening}

Decoder-LM embedding spaces are anisotropic; on our training captions the
mean pairwise cosine of raw frozen-text embeddings was 0.946 under the bare
input format. We standardize the frozen text side per dimension
(mean-center, divide by per-dimension standard deviation), fitted once from
cached caption embeddings and stored as buffers. The transform is
deliberately \emph{diagonal}: truncating to the first $D$ dimensions and
whitening with the first-$D$ statistics equals whitening then truncating,
so the Matryoshka nesting survives; full-covariance whitening would mix
dimensions and break it. Only the frozen text side is whitened; the
trainable audio side learns to match the whitened targets. Under the bare
input format whitening was worth +11.5 \ratk{10} (0.475 $\to$ 0.590 in a
4K-clip A/B). Under the native protocol the measured anisotropy drops
sharply (mean pairwise cosine 0.61--0.64; \S\ref{sec:anisotropy}), and
whitening's contribution shrinks to +3.3 \ratk{10} (0.593 $\to$ 0.626),
still real, so it stays in the recipe.

\subsection{Stage two: in-domain fine-tuning}
\label{sec:finetune}

End-to-end CLAP systems owe a documented +13--15 \ratk{1} to in-domain
AudioCaps fine-tuning after pretraining~\citep{bai2024audiosetcaps}. We add
the same stage under our constraint that only the trained components may
change. The fine-tune warm-starts the resampler (and in generation~2 the
adapters), the whitening buffers, and the temperature from the finished
pretraining checkpoint, keeps the \emph{pretraining} whitening transform
(the connector was trained against that exact target geometry; refitting on
the small in-domain corpus would move the targets under it), initializes a
fresh optimizer, and trains 400 steps on the AudioCaps train split alone,
in minutes of GPU time.

Two gates, pre-registered in the execution plan before the run, decide
acceptance: AudioCaps audio-to-text \ratk{10} must gain at least $+2$
points, and Clotho zero-shot audio-to-text \ratk{10} must lose no more
than $2$ points (a forgetting guard). Generation~1's accepted checkpoint
(v0.3) passes both: AudioCaps audio-to-text moves from 0.717/0.279 to
\textbf{0.741}/\textbf{0.332} (\ratk{10}/\ratk{1}; $+2.4$) and
text-to-audio \ratk{10} to 0.746; Clotho lands at 0.433 audio-to-text
($-1.5$, inside the gate) with text-to-audio \ratk{10} \emph{up} 1.1
points at 0.460. The cost is measurable but small elsewhere: emergent
audio$\to$image on VGGSound-696 gives back about one point (\ratk{10}
0.418 $\to$ 0.407); in-domain sharpening trades a little cross-modal
generality. The schedule length is at an \ratk{1}-preserving optimum, with
shorter and longer arms both worse (\S\ref{sec:ftlength}).

\subsection{Configuration and cost}
\label{sec:config}

The generation-1 headline run: 3{,}200 steps at effective batch 1024
(micro-batch 128 $\times$ 8 accumulation) $\approx$ 6.8 epochs over 484K
pairs, AdamW with cosine decay and 5\% linear warmup, bf16 base precision,
full-corpus negative bank, native-format targets, whitening on: 5.1
hours on a single H100 (peak 49.5\,GB), training loss 9.48 $\to$ 4.59,
\texttt{base\_drift} $= 0.0$. The
generation-2 pretrain is the same recipe on the 518K cleaned corpus
(\S\ref{sec:crowding}) with adapters enabled and 3{,}900 steps.
Audio-tower frames and caption embeddings are precomputed once (both
towers frozen) and streamed from shards, so every expensive forward pass is
paid exactly once.

\paragraph{Precision lineage as protocol.}
Train, fine-tune, and score run at one precision, end to end. We measured
that rescoring a bf16-trained checkpoint through a 4-bit base silently
costs 5.6 \ratk{10} points, and the trap is bidirectional (a 4-bit-trained
generation-2 lineage lost $\sim$2 points scored at bf16). Checkpoints
therefore record their base precision and the scorer warns on mismatch;
all release numbers come from matched bf16 lineages.

\section{Data and Scaling}
\label{sec:data}

\subsection{Corpus}
\label{sec:corpus}

The generation-1 headline (v0.2) corpus is 484{,}372 audio--caption pairs:
the full AudioCaps train split ($\sim$45K)~\citep{kim2019audiocaps},
FSD50K~\citep{fonseca2022fsd50k}, the WavCaps AudioSet-SL
subset~\citep{mei2023wavcaps}, and a 318K-clip subset of LAION-FreeSound.
All evaluation sets (AudioCaps test, Clotho, ESC-50, VGGSound) are excluded
from training by clip id at ingestion. Long clips are randomly cropped to
10\,s training windows (250 frames, the CLAP-standard window), which
bounds the per-clip cache volume and acts as light temporal augmentation.
The generation-2 corpus grows this to 592K pairs (a FreeSound tail and BBC
sound effects), then drops 73{,}716 clips whose metadata describes no sound
content, leaving 518{,}183 training pairs with their original captions.

\begin{figure}[t]
\centering
\includegraphics[width=\columnwidth]{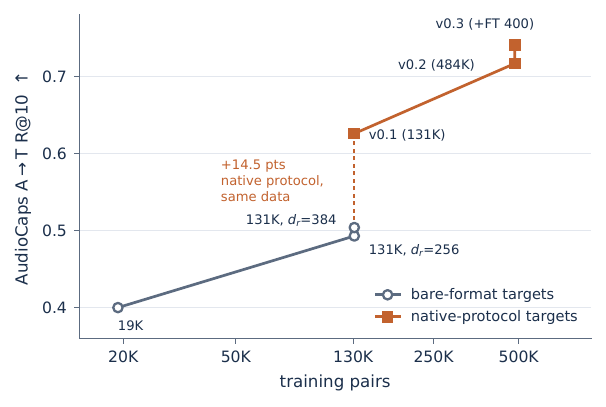}
\caption{AudioCaps A$\to$T \ratk{10} (883 clips, five-reference min-rank)
versus training-corpus size across the milestones of
Table~\ref{tab:ladder}. Grey circles: runs with bare-format text targets;
orange squares: native-protocol targets. The dotted riser marks the
+14.5-point native-protocol effect at \emph{fixed} data
(\S\ref{sec:native}); the 131K$\to$484K step adds a further +9.1 points,
and the 400-step in-domain stage reaches 0.741. Points differ in schedule
as labeled, so this is a milestone progression, not a controlled scaling
law.}
\label{fig:scaling}
\end{figure}

\begin{table*}[t]
\centering
\small
\begin{tabular}{llllcc}
\toprule
Stage & Corpus & Config & Change & \ratk{1} & \ratk{10} \\
\midrule
{[S0]} & 19K & bare, whiten, batch 128, 4000 steps & baseline & 0.100 & 0.400 \\
{[S1]} & 131K & + eff.\ batch 1024, 131K bank, $d_r{=}256$, 1800 steps & data + negatives & 0.134 & 0.493 \\
{[S2]} & 131K & $d_r{=}384$, 1800 steps & capacity & 0.143 & 0.504 \\
{[S3]} & 131K & $d_r{=}384$, 800 steps, \textbf{native targets} (v0.1) & protocol & 0.216 & 0.626 \\
 & 131K & native, \emph{without} whitening (control) & $-$whitening & 0.196 & 0.593 \\
{[S4]} & 484K & $d_r{=}384$, 3200 steps, native (v0.2) & data scale & 0.279 & 0.717 \\
{[S5]} & 484K & + 400-step in-domain FT (v0.3) & fine-tune stage & \textbf{0.332} & \textbf{0.741} \\
\midrule
\multicolumn{6}{l}{\emph{Generation 2}} \\
{[S6]} & 518K & + gated adapters $r{=}384$, soft-label/FN-mask, 3900 steps$^{\ddagger}$ & in-layer capacity & 0.268 & 0.708 \\
 & 592K & full corpus, no adapters, no loss terms (same-scale baseline)$^{\ddagger}$ & --- & 0.232 & 0.674 \\
{[S7]} & 518K & + 400-step in-domain FT (\feb{} release) & fine-tune stage & 0.302 & \textbf{0.743} \\
\bottomrule
\end{tabular}
\caption{The family recipe ladder on AudioCaps (A$\to$T). Each row changes
one axis relative to the row above unless noted; the [S2]$\to$[S3] step
changes schedule and target format together (its protocol component is
isolated in \S\ref{sec:native}). ``Bare''/``native'' refers to the
text-target input format. $^{\ddagger}$[S6] rows are the in-run probe
protocol at the jobs' 4-bit training precision (a matched pair,
comparable to each other but not to the release-protocol rows); [S7] is the
release protocol (\S\ref{sec:protocols}).}
\label{tab:ladder}
\end{table*}

\subsection{The ladder from 0.400 to 0.741}
\label{sec:ladder}

Table~\ref{tab:ladder} and Figure~\ref{fig:scaling} decompose the
generation-1 path. The individually largest levers were, in order: the
native-protocol targets (+14.5 \ratk{10} at fixed data, half schedule),
corpus scale (19K$\to$131K: +9.3; 131K$\to$484K with multi-epoch training:
+9.1 over v0.1), connector width at matched schedule (+4.7;
\S\ref{sec:capacity}), whitening (+11.5 bare / +3.3 native), the in-domain
fine-tuning stage (+2.4 \ratk{10}, +5.3 \ratk{1}; \S\ref{sec:finetune}),
and schedule length at matched width (+2.3). Notably, at 19K scale a
negatives-at-1024 arm did \emph{not} beat the plain 4000-step baseline
(\ratk{10} 0.392 vs.\ 0.400): large negative pools pay off only once
data scale stops being the binding constraint, consistent with the floor
audit below.

\subsection{Loss floor: crowding, not caption noise}
\label{sec:floor}

Training loss at 131K saturated at 4.0--5.0 from around step 400. A natural
hypothesis is a noise floor from semantically near-duplicate captions
(different clips, near-identical captions) acting as false negatives. We
tested it directly: re-run the exact training loss (same batch size
128, same 131K-row bank, same exact-duplicate masking, the checkpoint's
own temperature) with the audio embedding \emph{replaced by the whitened
text embedding}, i.e.\ a hypothetically perfect connector.

\begin{table}[t]
\centering
\small
\begin{tabular}{lc}
\toprule
Variant & Loss floor \\
\midrule
Observed config (131K bank + dup.\ mask) & 2.27 $\pm$ 0.11 \\
\quad + FN-mask bank at cos $\geq 0.95$ & 2.26 \\
\quad + FN-mask bank at cos $\geq 0.85$ & 2.15 \\
In-batch only (no bank) & 0.25 \\
\midrule
Observed saturated training loss & $\sim$4.0--5.0 \\
\bottomrule
\end{tabular}
\caption{Loss-floor audit at the exact 131K training configuration. A
perfect connector would reach 2.27; training saturates at 4.0--5.0, so
$\sim$2 nats of headroom are attributable to connector capacity /
optimization, not caption noise. False-negative masking of the bank moves
the floor by at most 0.12 nats even at an aggressively low threshold.}
\label{tab:floor}
\end{table}

Table~\ref{tab:floor}: the perfect-connector floor is \textbf{2.27},
roughly 2 nats below the observed saturation, so the saturation is
\emph{not} a caption-noise floor; capacity and optimization bind. Masking
near-duplicate bank entries buys only 0.01--0.12 nats: near-duplicates at
cosine $\geq$0.95 average just 5.4 per anchor. The bank's $\sim$2-nat floor
cost (2.27 vs.\ 0.25 without it) comes from \emph{crowding}, not twins: the
median nearest-neighbor cosine among 131K whitened caption embeddings is
0.914 (p90 0.966), i.e.\ the caption distribution has low effective
dimensionality ($\sim$30 by max-cosine statistics), so hundreds of
moderate-similarity negatives each leak a fraction of a positive's mass.
The audit also fixes a practical threshold: genuinely distinct captions
(``a woman gives a speech'' vs.\ ``a man gives a speech'') sit at cosine
0.976, so any relevance-based masking must use $\tau \gtrsim 0.98$. The
audit's verdict (widen the connector, then scale data) was
subsequently confirmed by intervention (\S\ref{sec:capacity} and the 484K
run).

\subsection{Scaling past 500K: bank crowding is real}
\label{sec:crowding}

Growing the corpus from 484K to 592K pairs at the fixed recipe
\emph{regressed} both benchmarks, the effect the floor audit predicts:
with a full-corpus bank, every added caption is also an added negative,
and the caption distribution's low effective dimensionality means added
mass lands close to existing anchors. The scale-up also changes corpus
composition (a FreeSound tail and BBC captions), so the regression is
\emph{consistent with} bank crowding rather than proof of it; the
intervention supports the diagnosis: in a matched A/B at 592K, enabling
the soft-label and false-negative-masking terms (\S\ref{sec:loss})
recovers +2.6 \ratk{10} (0.501 $\to$ 0.527, in-run probe protocol), and
these flags are adopted for all 500K+ runs. Corpus scale beyond
$\sim$500K pairs is therefore not free under a full-corpus negative bank.
The mechanism is the false-negative problem studied by debiased and
hard-negative contrastive
learning~\citep{chuang2020debiased,robinson2021hard}, surfacing here as a
bank-scaling caveat; AuroLA's soft-positive loss design independently
corroborates it~\citep{aurola2026}.

\section{Evaluation}
\label{sec:experiments}

\subsection{Protocols}
\label{sec:protocols}

\paragraph{AudioCaps.} We evaluate on 883 test clips with 4{,}411 reference
captions ($\sim$5 per clip), scoring audio-to-text as min-rank over the
five references (any reference in top-$k$ counts) and text-to-audio with
each caption as a query; this is the protocol used by published CLAP-family
results. Our 883-clip pool is the currently live-audio subset of the
$\sim$957-clip test convention; a larger candidate pool is, if anything,
harder. Because no single canonical split exists (papers variously use
957-, 816-, and private splits), we also built a machine-readable allowlist
of the 957-clip WavCaps/ACT convention (our shard covers 877/957) and
support scoring against it; on the 877-clip canonical subset an earlier
checkpoint scored \ratk{10} 0.503 vs.\ 0.504 on the full 883, indicating
the pool difference is immaterial. All retrieval is at the 1024-d
Matryoshka rung.

\paragraph{Clotho (zero-shot).} We use the canonical Clotho v2.1 evaluation
set~\citep{drossos2020clotho} built directly from the Zenodo release:
1{,}045 clips $\times$ exactly 5 captions, min-rank scoring. Clotho never
appears in our training data, so all Clotho numbers are strictly zero-shot;
we compare only against baselines that are also zero-shot on Clotho's
captions (most published Clotho rows are not).

\paragraph{Cross-modal.} VGGSound-696~\citep{chen2020vggsound} (696 audio
clips paired with video frames; chance \ratk{10} $= 0.014$). VGGSound is
blacklisted from our training data. Images pass through the base's native
vision path with its official embedding prompt; audio through the trained
audio path; text through the frozen text path. Rankings use per-modality
mean-centering of the gallery (\S\ref{sec:crossmodal}).

\paragraph{Paired-baseline replay.} Evaluation protocol can dominate the
measured number (\S\ref{sec:native}), and the sensitivity applies
\emph{longitudinally} within a project: a fresh evaluation may only be
compared against an archived result after replaying the archived run's
exact evaluation flags. During generation-2 development, a new
checkpoint's cross-modal score appeared to collapse relative to an
archived table; re-running the archived model as a paired baseline under
the fresh flags reproduced the same collapse, isolating the difference to
a template setting rather than the model. Our working practice, used for
every table in this section: score a paired baseline alongside every new
evaluation configuration before interpreting any cross-checkpoint
difference.

\paragraph{Uncertainty.} All results are single-seed. At the pool sizes
here a recall proportion carries a binomial standard error of roughly 1.5
points (883 queries) to 1.4 points (1{,}045), so we read single deltas
below $\sim$2 points as parity: the generation-2 AudioCaps audio-to-text
\ratk{10} edge over v0.3 (0.741 $\to$ 0.743) is within noise, while its
text-to-audio gains (+2.9 AudioCaps, +2.2 Clotho \ratk{10}) and its
VGGSound audio--text gains (+4.0/+3.6) exceed it. Bold in the tables marks
the best point estimate per column and class, not a significance claim.
No headline number uses test-time correction; a querybank-normalization
study is reported in Appendix~\ref{app:qbnorm}.

\subsection{Audio--text retrieval}
\label{sec:main-results}

\begin{table*}[t]
\centering
\small
\begin{tabular}{lccccc}
\toprule
 & & \multicolumn{2}{c}{A$\to$T} & \multicolumn{2}{c}{T$\to$A} \\
\cmidrule(lr){3-4}\cmidrule(lr){5-6}
Model & Trained params & \ratk{1} & \ratk{10} & \ratk{1} & \ratk{10} \\
\midrule
\multicolumn{6}{l}{\emph{Audio-native specialist systems (LoRA inside the backbone; audio--text only)}} \\
OEA~\citep{omniembedaudio2026} & 11--16M & --- & --- & 0.389 & 0.845 \\
AuroLA + re-ranker~\citep{aurola2026} & 418M + 7B re-ranker & \textbf{0.656} & \textbf{0.933} & \textbf{0.510} & \textbf{0.896} \\
\midrule
\multicolumn{6}{l}{\emph{CLAP-class dual encoders (both towers trained end-to-end + in-domain data)}} \\
LAION-CLAP~\citep{wu2023laionclap} & 0.15B & 0.468 & 0.907 & 0.361 & 0.839 \\
LAION-CLAP (released checkpoint, our protocol)$^{*}$ & 0.15B & 0.215 & 0.684 & 0.225 & 0.686 \\
WavCaps HTSAT-BERT~\citep{mei2023wavcaps} & $\sim$0.14B & 0.517 & 0.906 & 0.397 & 0.861 \\
Cacophony~\citep{zhu2024cacophony} & 0.21B & 0.553 & 0.924 & 0.410 & 0.864 \\
GLAP~\citep{dinkel2025glap} & 0.86B & 0.544 & 0.911 & 0.417 & 0.861 \\
M2D-CLAP~\citep{niizumi2025m2dclap} & $\sim$0.20B & \textbf{0.593} & \textbf{0.928} & \textbf{0.420} & \textbf{0.886} \\
\midrule
\multicolumn{6}{l}{\emph{Unified 3-modality space, all towers trained (fine-tuned on merged AudioCaps+Clotho+MACS train)}} \\
ONE-PEACE~\citep{wang2023onepeace} & 4B & 0.510 & 0.920 & 0.425 & 0.884 \\
\midrule
\multicolumn{6}{l}{\emph{Unified multi-modality embedders, frozen base (ours; text/image/video preserved bitwise)}} \\
\fea{} v0.1 (131K pairs) & 16.4M & 0.216 & 0.626 & 0.226 & 0.680 \\
\fea{} v0.2 (484K pairs) & 16.4M & 0.279 & 0.717 & 0.268 & 0.736 \\
\fea{} v0.3 (+ in-domain FT) & 16.4M & \textbf{0.332} & 0.741 & 0.280 & 0.746 \\
\feb{} (deep adapters) & 60.6M & 0.302 & \textbf{0.743} & \textbf{0.292} & \textbf{0.775} \\
\bottomrule
\end{tabular}
\caption{AudioCaps test retrieval, grouped by system class; bold marks the
best per column \emph{within each class}. Our rows: 883 clips,
five-reference min-rank protocol, bf16, native templates. All other rows
are the numbers reported by their authors (retrieved 2026-07-12) on their
respective AudioCaps test protocols, which vary in split and pool across
papers (\S\ref{sec:protocols}). $^{*}$Our reproduction: the public
\texttt{laion/clap-htsat-fused} checkpoint (pinned revision) scored on our
exact 883-clip five-reference protocol; the cited row above it is the
paper's own best AudioCaps configuration on its own protocol, so the two
rows differ in both checkpoint and protocol. M2D-CLAP's direction labels are
single-sourced (the reproduction in \citet{omniembedaudio2026} reports
T$\to$A \ratk{1} 0.414). AuroLA is a two-pass system (7B retriever, then a
second 7B pairwise re-ranker). ONE-PEACE trains all 4B parameters and
fine-tunes on the merged AudioCaps+Clotho+MACS train splits (its
evaluation split is not stated). Specialist and CLAP systems embed audio
and text only; our models additionally embed image and video in the same
space, with the base's performance there preserved bitwise
(\S\ref{sec:invariance}). Dashes: not reported / not recorded.}
\label{tab:audiocaps}
\end{table*}

\begin{table*}[t]
\centering
\small
\begin{tabular}{lccccc}
\toprule
 & & \multicolumn{2}{c}{A$\to$T} & \multicolumn{2}{c}{T$\to$A} \\
\cmidrule(lr){3-4}\cmidrule(lr){5-6}
Model & Trained params & \ratk{1} & \ratk{10} & \ratk{1} & \ratk{10} \\
\midrule
\multicolumn{6}{l}{\emph{Audio-native specialist systems}} \\
AuroLA (pretrain, ZS)~\citep{aurola2026} & 418M & \textbf{0.329} & --- & \textbf{0.265} & --- \\
\midrule
\multicolumn{6}{l}{\emph{CLAP-class dual encoders}} \\
WavCaps CNN14-BERT (ZS)~\citep{mei2023wavcaps} & --- & 0.217 & 0.576 & 0.175 & 0.549 \\
WavCaps HTSAT-BERT (ZS)~\citep{mei2023wavcaps} & $\sim$0.14B & 0.200 & 0.566 & 0.165 & 0.509 \\
GLAP~\citep{dinkel2025glap} & 0.86B & \textbf{0.218} & \textbf{0.615} & \textbf{0.194} & \textbf{0.583} \\
\midrule
\multicolumn{6}{l}{\emph{Unified multi-modality embedders, frozen base (ours)}} \\
\fea{} v0.1 (131K) & 16.4M & 0.064 & 0.252 & 0.085 & 0.329 \\
\fea{} v0.2 (484K) & 16.4M & \textbf{0.135} & \textbf{0.448} & 0.136 & 0.449 \\
\fea{} v0.3 (+ AudioCaps FT) & 16.4M & \textbf{0.135} & 0.433 & 0.136 & 0.460 \\
\feb{} (deep adapters) & 60.6M & 0.127 & 0.421 & \textbf{0.151} & \textbf{0.482} \\
\bottomrule
\end{tabular}
\caption{Clotho v2.1 evaluation (1{,}045 clips $\times$ 5 references),
zero-shot rows only: every listed model excludes Clotho's \emph{caption}
data from training; bold: best per column within each class. A caveat
applies to the baselines' audio: WavCaps' FreeSound portion overlaps 61\%
of the Clotho evaluation audio (as measured by
\citet{omniembedaudio2026}), so zero-shot rows for WavCaps-trained models
(including GLAP, which trains on WavCaps) are optimistic; AuroLA's
training-audio overlap with Clotho is not reported. Our corpus excludes
all 5{,}929 Clotho FreeSound clip ids by blacklist at ingestion. Baseline
numbers as reported by their authors (retrieved 2026-07-12); GLAP numbers
are single-sourced. Dashes: not reported / not recorded.}
\label{tab:clotho}
\end{table*}

Table~\ref{tab:audiocaps} reports AudioCaps retrieval across the system
classes; Table~\ref{tab:clotho} the strictly zero-shot Clotho comparison.
Consistent with \S\ref{sec:native}, the cited rows ride their authors'
input protocols and splits, so we read these tables at the class level
rather than cell against cell. Within the frozen-base class,
generation~1's pretraining checkpoint (v0.2) reaches audio-to-text
\ratk{10} 0.717 (\ratk{1} 0.279, \ratk{5} 0.590, mAP@10 0.208); the
in-domain stage (v0.3) lifts this to 0.741/0.332, and generation~2 posts
the family's best text-to-audio results on both benchmarks (AudioCaps
0.775, Clotho 0.482) while matching v0.3 on audio-to-text \ratk{10}
(0.743, within noise; \S\ref{sec:protocols}); full \ratk{1}/\ratk{5}/%
\ratk{10} grids are in Appendix~\ref{app:grids}. The remaining gap to the
fully fine-tuned CLAP family is 16--19 \ratk{10} points, and the
audio-native specialist systems sit higher still; these are comparisons whose
structural terms (both towers trained, in-domain data, LoRA inside the
backbone, a second re-ranking pass) are drawn in the table and taken up
in \S\ref{sec:limitations}. As a protocol anchor, we also scored the
released public LAION-CLAP checkpoint on our exact protocol (the starred
row): audio-to-text \ratk{10} 0.684, below both v0.2 and v0.3.
The gap to the cited row is therefore a cross-protocol, cross-checkpoint
reading, the effect \S\ref{sec:native} documents, and the tables support
class-level conclusions rather than cell-level ones. On Clotho, our models sit within 10 \ratk{10}
points text-to-audio (17 audio-to-text) of the strongest nominally
zero-shot dual encoder despite frozen towers, with the caveat in the
caption that those baselines' training audio overlaps the Clotho
evaluation set while ours is excluded by construction; ONE-PEACE's Clotho
rows (audio-to-text 0.271/0.654, text-to-audio 0.224/0.627
\ratk{1}/\ratk{10}) are omitted from Table~\ref{tab:clotho} because its
training includes Clotho's train split. The \ratk{10}/\ratk{1} ratio, a
measure of top-of-ranking sharpness (lower is sharper), improved from
2.9$\times$ (v0.1) to 2.57$\times$ (v0.2) to 2.23$\times$ (v0.3);
end-to-end systems sit at 1.7--2.2$\times$.

\paragraph{The family delta.}
Table~\ref{tab:fe2} compares the two generations cell by cell under the
release protocol. The generation-2 gains concentrate in the text-to-audio
direction and the cross-modal audio--text pair ($+4.0$/$+3.6$ \ratk{10},
$+5.3$ T$\to$A \ratk{1} on VGGSound); generation~1 retains audio-to-text
\ratk{1} on both caption benchmarks. The pattern is consistent with the
mechanism: in-layer capacity lets the audio representation organize for
discrimination against text queries (the direction that improves
everywhere), at a small cost in audio-to-text top-1 on the caption benchmarks.
In the controlled A/B behind this comparison (matched 45K-scale arms, 800
steps, identical recipe; Appendix~\ref{app:probe}), adapters at rank 384
improve every retrieval
direction (audio-to-text \ratk{10} 0.631 $\to$ 0.665 ($+3.4$), \ratk{1}
0.202 $\to$ 0.229, text-to-audio \ratk{10} 0.684 $\to$ 0.708), with rank
128 recovering $+2.5$, so the gains scale with adapter capacity and had not
saturated at the largest rank trained. The experiment replicates across
three seeds, with paired audio-to-text \ratk{10} deltas of
$+3.4$/$+4.6$/$+5.4$ and every retrieval direction positive at every seed;
the published pairing is the most conservative of the three. At full 518K
scale the pretraining gain reproduces (\ratk{10} 0.708 vs.\ 0.674, matched
protocol).

\begin{table}[t]
\centering
\small
\begin{tabular}{lcc}
\toprule
 & Gen.~1 (v0.3) & Gen.~2 \\
\midrule
AudioCaps A$\to$T \ratk{1} & \textbf{0.332} & 0.302 \\
AudioCaps A$\to$T \ratk{10} & 0.741 & \textbf{0.743} \\
AudioCaps T$\to$A \ratk{1} & 0.280 & \textbf{0.292} \\
AudioCaps T$\to$A \ratk{10} & 0.746 & \textbf{0.775} \\
Clotho A$\to$T \ratk{1} & \textbf{0.135} & 0.127 \\
Clotho A$\to$T \ratk{10} & \textbf{0.433} & 0.421 \\
Clotho T$\to$A \ratk{1} & 0.136 & \textbf{0.151} \\
Clotho T$\to$A \ratk{10} & 0.460 & \textbf{0.482} \\
VGGSound A$\to$T \ratk{10} & 0.625 & \textbf{0.665} \\
VGGSound T$\to$A \ratk{1} & 0.213 & \textbf{0.266} \\
VGGSound T$\to$A \ratk{10} & 0.645 & \textbf{0.681} \\
Emergent A$\to$I \ratk{10} & \textbf{0.407} & 0.392 \\
\bottomrule
\end{tabular}
\caption{The family delta under the release protocol (bf16, native
templates; Clotho zero-shot). Bold marks the better generation per row
(deltas under $\sim$2 points are within single-seed noise;
\S\ref{sec:protocols}). The pre-fine-tune generation-2 checkpoint reaches
the family's best emergent audio-to-image alignment (\ratk{10} 0.443,
measured under the earlier readout template), indicating the adapters
strengthen rather than erode the text bridge; the in-domain fine-tune
then gives back part of that alignment for its AudioCaps gains (0.392
under the release readout; the two numbers straddle a template change,
so the exact size of the trade is not protocol-clean), as the
generation-1 fine-tune did ($-1.1$).}
\label{tab:fe2}
\end{table}

\subsection{Breadth: MAEB, user-intent queries, and coverage}
\label{sec:breadth}

\paragraph{MAEB.} We evaluated the released generation-1 v0.2-preview
checkpoint with \texttt{mteb==2.18.0} on the sound-event tier of the MAEB
benchmark~\citep{maeb2026}: nine task results submitted to the public
leaderboard (BeijingOpera, ClothoT2ARetrieval, GTZANAudioReranking,
GTZANGenre, MACST2ARetrieval, RavdessZeroshot,
SpeechCommandsZeroshotv0.02, UrbanSound8KT2ARetrieval,
VehicleSoundClustering); against the public leaderboard as of 2026-07-09
these scores place roughly \#3--\#6 per task on the tasks
closest to our training domain (UrbanSound8K \#3, Ravdess \#4) despite the
speech- and music-light corpus;
per-task scores are in Appendix~\ref{app:maeb}. A tenth task,
FSD2019Kaggle, was evaluated
but withheld: matching its test clips by FreeSound id against the Zenodo
post-competition metadata shows 608 of its 4{,}481 test clips (13.6\%)
appear in the FSD50K development split (part of our training corpus),
and the test set is in fact entirely contained in FSD50K. Any
FSD50K-trained model is affected on that task; we flagged this to the
benchmark maintainers.

\paragraph{User-intent queries.} \citet{omniembedaudio2026} argue that
caption-style queries understate real search behavior and introduce
user-intent queries: questions, commands, keyword tags, paraphrases, and
exclusion queries. On our own evaluation in that style over the identical
1{,}045-clip Clotho pool, v0.3 averages \ratk{10} 0.518 across the four
positive query formulations: retrieval degrades gracefully rather than
collapsing off the caption manifold, which we attribute to the LLM
backbone's instruction conditioning.

\begin{figure*}[t]
\centering
\includegraphics[width=\textwidth]{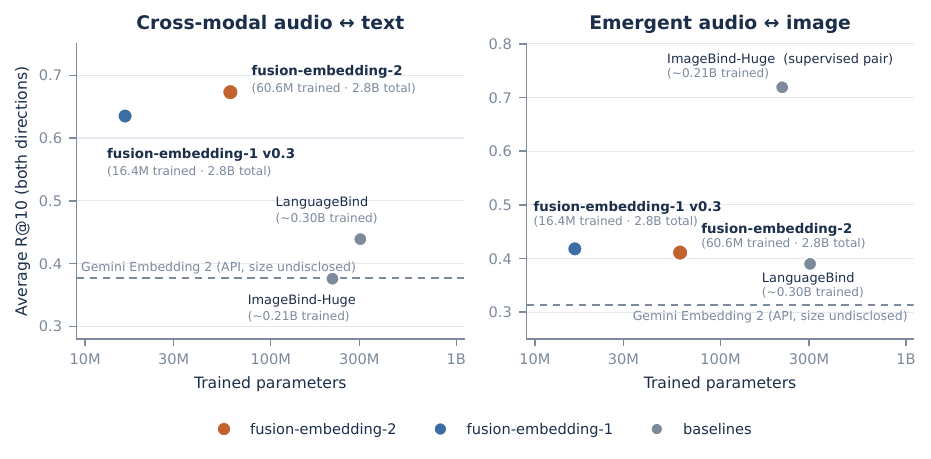}
\caption{Positioning among unified embedding models on VGGSound-696
cross-modal retrieval (\ratk{10} averaged over both directions; log axis
of \emph{trained} parameters). Left: audio$\leftrightarrow$text. Right:
emergent audio$\leftrightarrow$image, where ImageBind trains on its
supervised audio--image pair while our models have seen zero audio--image
pairs. Trained-parameter positions: \fea{} 16.4M, \feb{} 60.6M, ImageBind
$\approx$0.21B trained of 1.2B total (frozen CLIP towers), LanguageBind
$\approx$0.30B (the fully fine-tuned audio tower we benchmark). Baselines
are our reproductions on the identical protocol (\S\ref{sec:crossmodal});
the dashed line is Gemini Embedding~2, an API of undisclosed size,
evaluated 2026-07-09.}
\label{fig:positioning}
\end{figure*}

\paragraph{Coverage.} The evaluation surface itself separates the system
classes, and Table~\ref{tab:coverage} makes the claim explicit. Every
specialist row in Tables~\ref{tab:audiocaps} and~\ref{tab:clotho} embeds
audio and text only: none of those systems can populate a single cell of
the cross-modal matrix in \S\ref{sec:crossmodal}, and none inherits a
text/image/video benchmark result at all. Our models populate every cell
of every table in this paper with one checkpoint and one space, and the
text/image/video cells of that space are the base model's published
MMEB-V2 results~\citep{qwen3vlembedding}, carried over bitwise
(\S\ref{sec:invariance}). Figure~\ref{fig:positioning} places the family
among the systems that \emph{can} enter the unified comparison.

\begin{table*}[t]
\centering
\footnotesize
\begin{tabular}{lccccccc}
\toprule
 & Trained & \multicolumn{4}{c}{Base benchmarks (0--100)} & \multicolumn{2}{c}{Audio, this work (\ratk{10})} \\
\cmidrule(lr){3-6}\cmidrule(lr){7-8}
System & params & Text & Image & Video & VisDoc & AC A$\to$T & VGG a$\leftrightarrow$i \\
\midrule
\fea{} v0.3 & 16.4M & 63.9$^{\circ}$ & 75.0$^{\circ}$ & 61.9$^{\circ}$ & 79.2$^{\circ}$ & 0.741 & 0.418 \\
\feb{} & 60.6M & 63.9$^{\circ}$ & 75.0$^{\circ}$ & 61.9$^{\circ}$ & 79.2$^{\circ}$ & 0.743 & 0.411 \\
OEA~\citep{omniembedaudio2026} & 11--16M & --- & --- & --- & --- & --- & --- \\
AuroLA~\citep{aurola2026} & 418M + 7B & --- & --- & --- & --- & 0.933 & --- \\
GLAP~\citep{dinkel2025glap} & 0.86B & --- & --- & --- & --- & 0.911 & --- \\
ONE-PEACE~\citep{wang2023onepeace} & 4B & --- & --- & --- & --- & 0.920 & --- \\
ImageBind-Huge~\citep{girdhar2023imagebind} & $\sim$0.21B & --- & --- & --- & --- & --- & 0.719$^{*}$ \\
LanguageBind~\citep{zhu2024languagebind} & $\sim$0.30B & --- & --- & --- & --- & --- & 0.390$^{*}$ \\
\bottomrule
\end{tabular}
\caption{Coverage across the unified evaluation surface. Cells are each
system's published results on the named benchmark, or our reproductions
($^{*}$, VGGSound-696 average \ratk{10} over both directions on the
protocol of \S\ref{sec:crossmodal}); base-benchmark cells are official
0--100 scores (Text $=$ MMTEB Mean(Task); Image/Video/VisDoc $=$ MMEB-V2
category overalls), audio cells are recall fractions. $^{\circ}$ marks
cells inherited from the released Qwen3-VL-Embedding-2B
card~\citep{qwen3vlembedding} bitwise (\S\ref{sec:invariance}): not
re-measured, mechanically preserved. The table's claim is
\emph{population}, not cell-vs-cell comparison: one \family{} checkpoint
covers the entire row; every other system leaves most of the surface
empty. ImageBind's audio--image cell is its supervised training pair.}
\label{tab:coverage}
\end{table*}

\subsection{Invariance verification}
\label{sec:invariance}

The guarantee of \S\ref{sec:guarantee} is enforced, not assumed, and every
check reads exactly zero. \emph{Parameters}: the regression guard's
\texttt{base\_drift} assertion, the maximum absolute change over every
base parameter, returned $0.0$ on every training run reported in this
paper, including all generation-2 adapter runs. \emph{Outputs}: unit tests
encode identical text through the base model and through \feb{} with
adapters attached and assert bitwise-equal activations (maximum absolute
difference $0$, not merely small); the release smoke test repeats the check
on the packaged checkpoint through the public loading path. \emph{Controls
that must differ, do}: the same test suite verifies that audio-gated
forwards \emph{change} when adapters are enabled, that gradients reach only
adapter and connector parameters, and that a gate held open across an
interrupted backward is detected loudly rather than silently degrading the
guarantee. Text--image retrieval cells for \fea{} and \feb{} in
Table~\ref{tab:crossmodal} are identical to the third decimal because the
underlying vectors are identical to the last bit. Table~\ref{tab:invariance}
summarizes the checks; the full verification protocol, mapped to the
open-source test suite, is in Appendix~\ref{app:invproto}.

\begin{table}[t]
\centering
\small
\begin{tabular}{lcc}
\toprule
Check & Gen.~1 & Gen.~2 \\
\midrule
Base param.\ drift, max $|\Delta|$, every run & 0 & 0 \\
Text forward $\Delta$ vs.\ base (unit test) & 0 & 0 \\
Text forward $\Delta$ (released ckpt.,\ smoke) & 0 & 0 \\
Text$\leftrightarrow$image vectors across family & \multicolumn{2}{c}{identical} \\
\bottomrule
\end{tabular}
\caption{Invariance verification: every entry is a maximum absolute
difference, exactly zero (not approximately) under matched execution
(\S\ref{sec:guarantee}). Image and video inputs take the same gated path
as text (the gate keys on audio presence, not token type), so the text
checks cover the mechanism; the identity of every
text$\leftrightarrow$image cell across all four released checkpoints
(Table~\ref{tab:crossmodal}) confirms it end-to-end.}
\label{tab:invariance}
\end{table}

\section{Emergent Cross-Modal Retrieval}
\label{sec:crossmodal}

The connector never sees an audio--image pair; VGGSound itself is
blacklisted from training. \textbf{Because we train no audio--visual
supervision of any kind, we call retrieval between audio and the base's
visual modalities \emph{emergent cross-modal retrieval}}, the retrieval
analogue of the emergent zero-shot classification that ImageBind
established for bind-to-pivot training~\citep{girdhar2023imagebind}, here
arising from a frozen base whose text--image geometry the new modality
simply inherits. The setting is strictly harder than supervised
cross-modal retrieval, and published baselines for it are scarce:
ImageBind is the natural reference point, with its supervised
audio--image pair printed for calibration rather than competition.

Generation~1 v0.2 retrieves the matching video frame from sound alone at
\ratk{10} 0.418 (\ratk{1} 0.088, \ratk{5} 0.315, mAP@10 0.179),
29$\times$ chance, and the reverse direction reaches 0.440
(31$\times$). An earlier checkpoint, before the native-protocol change,
had already established the existence proof at \ratk{10} 0.310
(21$\times$ chance); audio--text quality improvements have since lifted
the emergent direction in lockstep: benchmark gains on the supervised
pair compound directly into the unsupervised ones. The strongest emergent
checkpoint in the family is the generation-2 pretrain at \ratk{10} 0.443.

\begin{table*}[t]
\centering
\small
\begin{tabular}{lccc}
\toprule
Model & audio$\leftrightarrow$image & audio$\leftrightarrow$text & text$\leftrightarrow$image \\
\midrule
Chance & 0.014 & 0.014 & 0.014 \\
ImageBind-Huge~\citep{girdhar2023imagebind} & \textbf{0.718 / 0.720} & 0.404 / 0.348 & 0.243 / 0.282 \\
LanguageBind~\citep{zhu2024languagebind} & 0.365 / 0.415 & 0.547 / 0.331 & 0.221 / 0.283 \\
Gemini Embedding 2~\citep{gemini2026embedding} & 0.312 / 0.316 & 0.379 / 0.374 & 0.273 / \textbf{0.366} \\
\fea{} v0.1 & 0.368 / 0.388 & 0.555 / 0.592 & \textbf{0.331} / 0.319 \\
\fea{} v0.2 & 0.418 / 0.440 & 0.588 / 0.631 & \textbf{0.331} / 0.319 \\
\fea{} v0.3 & 0.407 / 0.428 & 0.625 / 0.645 & \textbf{0.331} / 0.319 \\
\feb{} & 0.392 / 0.430 & \textbf{0.665 / 0.681} & \textbf{0.331} / 0.319 \\
\bottomrule
\end{tabular}
\caption{VGGSound-696 cross-modal retrieval, \ratk{10} in both directions
(chance $=0.014$). Identical clips, frames, and scoring for all models;
ImageBind numbers computed with the released \texttt{imagebind\_huge}
checkpoint, LanguageBind with revision-pinned \texttt{Audio\_FT} and
\texttt{Image} checkpoints (its audio--text shown with its strongest
readout, the audio branch's own text tower); our readout uses per-modality
mean-centering. ImageBind trains directly on audio--image pairs (its
supervised pair); LanguageBind trains audio against language; Gemini
Embedding~2 is a commercial natively-multimodal API (training
undisclosed), evaluated 2026-07-09 at its documented default invocation;
our models train on audio--text only; each model's other pairs are
emergent. Our text$\leftrightarrow$image rows are identical across the
family because text and image never touch the trained components
(\S\ref{sec:invariance}). Evaluation captions are the model-generated
descriptions distributed with the MAEB VGGSound retrieval
set~\citep{maeb2026}; the generating model is not documented by the
benchmark, and model-generated caption style could favor text towers
trained on similar registers (a sensitivity \S\ref{sec:native} shows
can be large), but all models received identical inputs. Bold: best per
pair (per direction where split).}
\label{tab:crossmodal}
\end{table*}

\begin{figure*}[tp]
\centering
\includegraphics[width=0.8\textwidth]{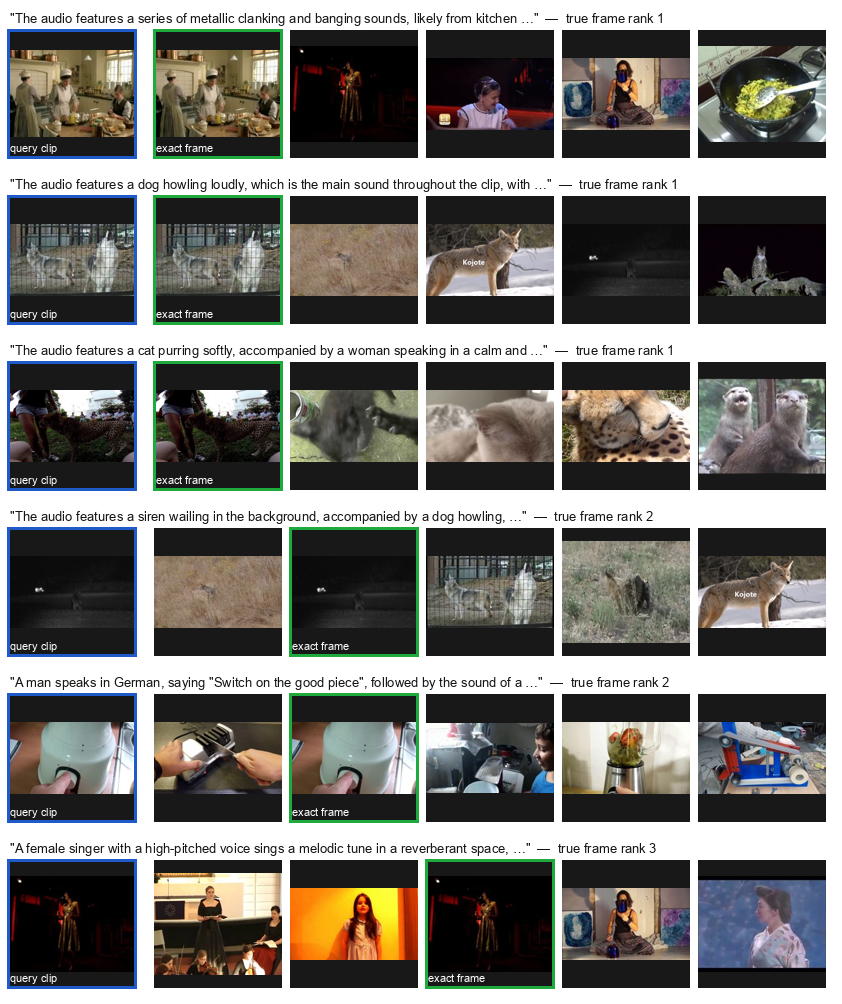}
\caption{Qualitative audio$\to$image retrieval on VGGSound-696 (v0.2
checkpoint; aggregate \ratk{10} 0.418, chance 0.014, zero audio--image
training pairs). Each row: the query clip's own frame (left, blue), then
the top-5 retrieved frames; green marks the clip's exact frame, with its
rank noted above each row. Retrievals are organized by sound category even
when the exact frame ranks lower. Example frames from the VGGSound
dataset~\citep{chen2020vggsound} (CC-BY-4.0), shown for evaluation
illustration.}
\label{fig:gallery}
\end{figure*}

\paragraph{Head-to-heads.}
Table~\ref{tab:crossmodal} reports the comparison on our exact protocol.
Against ImageBind-Huge, each model wins its \emph{supervised} pair:
ImageBind trains on audio--image (0.718/0.720 vs.\ our emergent
0.418/0.440); we train on audio--text (0.588/0.631 for v0.2, 0.665/0.681
for \feb, vs.\ its emergent 0.404/0.348). Comparing emergent-vs-emergent,
ImageBind's audio--text (0.404/0.348) and our audio--image (0.418/0.440)
are of the same order. Our models additionally win text--image
(0.331/0.319 vs.\ 0.243/0.282), where both models are exercising
pretrained vision--language alignment; two of three pairs on this protocol
go to the frozen-base models.

Against LanguageBind the comparison is uniform: v0.2 exceeds it on all
three pairs, including LanguageBind's own supervised direction
(audio--text: 0.588/0.631 vs.\ 0.547/0.331 under its strongest readout),
and \feb{} extends the margin. A measurement on the released checkpoints
suggests why. LanguageBind's modality branches each fine-tune their own
copy of the text tower, and the copies diverge: the mean cosine between
the audio branch's and the image branch's embeddings of identical captions
is 0.553, so the language pivot intended to bind the branches is no longer
a shared space, and its emergent audio--image (0.365/0.415) falls below
ours despite a far larger trained parameter count. This failure mode is
structurally unavailable to a frozen-base design: every modality of the
\family{} family conforms to one immutable space, asserted
parameter-identical after every run.

Gemini Embedding~2~\citep{gemini2026embedding}, a commercial
natively-multimodal embedding API released in 2026, provides the closest
product-level analogue to our design goal (one space for text, images,
video, audio, and documents). On this protocol our models lead it on
audio--image in both directions (our \emph{emergent} pair against a
natively-trained system) and on audio--text in both directions, while
text--image splits by direction. We evaluate the API at its documented
default invocation and report the evaluation date; as a served model it
may change over time, and its training data and protocol are undisclosed,
so this row is a product snapshot rather than a method comparison.

\paragraph{Why alignment emerges.}
The mechanism is the one ImageBind established, inverted in an
economically useful direction. The base model already places text, images,
and video in one space. Our audio path is optimized solely to place audio
near its captions \emph{in that space}, and captions of a barking dog
live near images of barking dogs by the base's own training. Alignment to
text is therefore alignment to everything text is aligned to.
Quantitatively, the emergent audio$\leftrightarrow$image direction reaches
roughly 70\% of the strength of the trained audio$\leftrightarrow$text
direction on the same clips (0.418/0.440 vs.\ 0.588/0.631), and
improvements to the trained direction have transferred to the emergent one
at every checkpoint measured (0.368/0.388 $\to$ 0.418/0.440 from v0.1 to
v0.2, with the text$\leftrightarrow$image control unchanged at 0.331/0.319),
confirming the gain is attributable to audio moving, not to any change
in the base's geometry, which is impossible here by construction.
Figure~\ref{fig:gallery} shows the emergent direction qualitatively:
retrievals organize by sound category, with the query clip's exact frame
frequently in the top ranks.

\paragraph{Geometry: anisotropy is substantially a protocol artifact.}
\label{sec:anisotropy}
Under the bare input format, the frozen text embeddings of random training
captions have mean pairwise cosine 0.946 (measured at whitening-fit time;
0.93--0.95 across fits), the ``severe anisotropy'' of decoder-LM
embedding spaces. Under the native chat template the same statistic drops
to 0.640 (131K corpus) and 0.612 (484K corpus). Much of the reported
anisotropy of decoder-LM embedding spaces, in this model at least, is the
cost of querying the model off its training distribution rather than an
intrinsic property of the space. This is consistent with the whitening
ablation (\S\ref{sec:whitening}): the correction is worth +11.5 \ratk{10}
where the anisotropy is severe (bare) and +3.3 where it is mild (native).
On the cross-modal benchmark, readout geometry is a second-order effect:
per-modality mean-centering of the gallery is the best readout
(audio$\to$image 0.397 raw, 0.402 whitened, 0.418 centered for v0.2), and
the mismatch between the whitened text space and the raw image space
costs only $\sim$1 point.

\section{What Did Not Work}
\label{sec:negative}

Three controlled experiments refuted the three most natural improvement
hypotheses for this architecture. We report them at the same standard as
the positive results, each with its experiment, its number, and the
mechanism we believe explains it, because together they redraw the
design map: the binding constraint of frozen-backbone audio fusion is not
the captions, not the audio encoder, and not the connector, but the frozen
LM's ability to process audio tokens. Generation~2 (\S\ref{sec:adapters-arch})
is the direct response, and its gains land exactly where this section
predicts.

\subsection{Do cleaner captions help?}
\label{sec:recaption}

No: style dominates cleanliness. Roughly three quarters of our corpus
carries metadata-derived captions (titles, tags, recording notes) rather
than descriptive sentences, an obvious target for model-based
rewriting, which the caption-quality literature (\S\ref{sec:related})
would predict to help. We rewrote all 426K FreeSound and BBC captions with
an instruction-following LLM into short descriptive caption style (median
7 words), dropped the 74K clips whose metadata carried no sound content at
all, and retrained the full recipe on the remaining 518K pairs with the
large-scale loss terms enabled. The rewritten corpus is measurably cleaner
(exact-duplicate caption rows fall from 50.7\% to 45.0\%), yet
in-domain retrieval \emph{drops}: AudioCaps \ratk{10} 0.649 against 0.674
for the same-scale baseline trained on the raw captions (592K before the
junk-clip removal, without the loss terms; all arms in
Appendix~Table~\ref{tab:recaparms}). The $-2.5$ is a \emph{lower bound}
on the rewrite's cost there: the loss terms alone are worth $+2.6$ in a
matched A/B (\S\ref{sec:crowding}) and the junk-clip removal is
caption-preserving, so both differences between the arms favor the
rewritten one. (Measured against the 484K v0.2 checkpoint the drop reads
$-6.8$, but most of that is the scale regression of \S\ref{sec:crowding},
not the captions. Clotho reads 0.438, above the same-scale baseline's
0.400 and below v0.2's 0.448, consistent with the style analysis below
rather than with caption quality.)

The mechanism is a style-region shift, not a failure to learn: the run's
internal held-out evaluation \emph{on rewritten captions} reaches
\ratk{10} 0.84, while evaluation on original-style captions collapses, and
the run's whitening-fit statistics shifted sharply away from the
raw-caption fit, recording the text-distribution move directly. The
rewrite also targeted the wrong register for the benchmark it was meant to
help: the outputs are short AudioCaps-like sentences, where Clotho's
references are long, elaborate descriptions. The conclusion: \emph{caption
style specificity dominates caption cleanliness} for retrieval transfer;
rewriting a corpus into a generic clean style moves the model's operating
point rather than improving it. The fix implied is style-targeted:
match the reference register of the benchmark, and augment rather than
replace original captions. (The 74K-clip junk removal, by contrast, was
kept: it is caption-preserving and defines the generation-2 corpus.)

\subsection{Is a leaderboard-stronger audio encoder a better tower?}
\label{sec:towerswap}

No: co-training compatibility dominates encoder ranking. GLAP's controlled
ablation~\citep{dinkel2025glap} trains five frozen audio
encoders under one contrastive recipe with MLP projections and finds
Whisper-family encoders $\sim$9--12 mAP@10 behind sound-event encoders on
sound retrieval (AudioCaps T2A mAP@10: CED-Base 58.6, BEATs 55.1, Dasheng
55.8, Whisper-Base 46.5). Our tower is Whisper-family, so we ran the
implied experiment: a matched A/B at 45K AudioCaps-only scale, 800 steps,
identical recipe and hyperparameters, each arm evaluated on frames from
its own tower. The ranking \emph{inverts}: the Qwen2.5-Omni tower
(Whisper-family, $\sim$640M, co-trained to feed a Qwen LM) reaches
audio-to-text \ratk{10} 0.631 against 0.469 for
Dasheng-base~\citep{dinkel2024dasheng}, a 16-point margin, consistent
in both directions (text-to-audio 0.684 vs.\ 0.538). We note the size
confound: the towers are not parameter-matched (86M vs.\ $\sim$640M), so
the experiment does not isolate co-training from scale. It does establish
the practical point: encoder-quality rankings measured under shallow
projections do not transfer to architectures that splice tokens into a
frozen LLM, where compatibility with the LLM's expected input distribution
dominates. Selecting a tower for this architecture class by CLAP-style
leaderboards would have selected the wrong tower.

Two adjacent probes complete the picture. Whisper-family encoders are
reported to hide sound-event information in intermediate layers, with
last-layer features specializing toward speech~\citep{gong2023whisperat};
we linear-probed the actual frozen tower on
ESC-50~\citep{piczak2015esc50} (5-fold) and found the layer curve rises
\emph{monotonically} to 0.924 at the last layer, with a learned weighted
average over all layers \emph{worse} (0.893); the Omni tower's
continued pretraining evidently repaired the vanilla-Whisper pathology,
so multi-layer taps offer nothing for this tower. And the only controlled
same-connector/same-LLM ablation we are aware
of~\citep{interspeech2024cedwhisper} finds adapting a Whisper-family
encoder buys nothing for audio captioning (SPIDEr-FL: LoRA 44.8 vs.\
frozen 44.8, with full fine-tuning \emph{worse} at 44.0, versus 49.6 for
a frozen event encoder); adaptation capacity is not what frozen
Whisper-derived features lack. Together these results rule out
audio-tower adaptation as a productive direction for this architecture.

\subsection{Is the connector too small?}
\label{sec:capacity}

\begin{table}[t]
\centering
\small
\begin{tabular}{lcc}
\toprule
 & \ratk{1} & \ratk{10} \\
\midrule
\multicolumn{3}{l}{\emph{131K pairs, matched 800-step arms}} \\
$d_r=256$ & 0.096 & 0.434 \\
$d_r=384$ & \textbf{0.125} & \textbf{0.481} \\
$d_r=512$ & 0.119 & 0.479 \\
\midrule
\multicolumn{3}{l}{\emph{484K pairs, matched 3200-step runs}} \\
$d_r=384$ (v0.2) & \textbf{0.279} & \textbf{0.717} \\
$d_r=512$ & 0.249 & 0.675 \\
\bottomrule
\end{tabular}
\caption{Connector-width study on AudioCaps-883 (A$\to$T). In both
groups $d_r{=}512$ shows better \emph{training} metrics but equal or worse
held-out retrieval.}
\label{tab:capacity}
\end{table}

No: wider connectors memorize. At 131K with matched 800-step schedules
(Table~\ref{tab:capacity}),
$d_r{=}384$ beats $d_r{=}256$ by +4.7 \ratk{10} points (+30\% \ratk{1})
and reaches parity with the 256-width 1800-step run (0.481 vs.\ 0.493,
within single-seed noise; \S\ref{sec:protocols}) at 44\% of the steps.
Width 512
ties 384 on held-out retrieval \emph{despite} far better training metrics
(tail in-batch accuracy $\sim$0.73 vs.\ $\sim$0.53 and a much stronger
in-domain probe): the extra width fits the training distribution without
generalizing. The re-test at 484K confirms the knee is not an artifact of
the smaller corpus: an exact v0.2 mirror at $d_r{=}512$ loses on every
metric (\ratk{10} 0.675 vs.\ 0.717) with the same memorization signature.
We fix $d_r{=}384$. Input-side width, like the input-side tower, is not
where the headroom lives: capacity helps only when placed \emph{inside}
the layers, behind the modality gate (\S\ref{sec:main-results}).

\paragraph{Fine-tune length.}
\label{sec:ftlength}
The fine-tuning schedule shows the same shape in miniature: a 600-step
generation-1 arm improves AudioCaps \ratk{10} to 0.750 and text-to-audio
\ratk{10} to 0.762, with Clotho at 0.442/0.477, but AudioCaps \ratk{1}
falls from 0.332 to 0.316; a 200-step generation-2 arm is worse than 400
on both metrics (0.273/0.695 vs.\ 0.302/0.743). The two
arms sit on different generations, so they are consistent with---not
proof of---a shared optimum at 400 steps; we report 400-step
checkpoints throughout.

\section{Discussion and Limitations}
\label{sec:limitations}

On in-domain audio--text retrieval, the specialist systems remain ahead:
the CLAP family reaches AudioCaps audio-to-text \ratk{10} 0.906--0.928 by
training both towers end-to-end with in-domain data, and AuroLA
unsurprisingly achieves the best results of any system, but it trains
418M parameters inside a 7B backbone, adds a second 7B model as a
re-ranking pass, and, like every audio--text system in
Table~\ref{tab:audiocaps}, serves exactly one modality pair with no
preservation claim of any kind. The trade this family makes is explicit:
some in-domain audio--text accuracy, in exchange for one space over four
modalities, a bitwise-preserved base, single-GPU-hours training cost, and
an index that never needs re-embedding.

Our own scale-up evidence tempers the obvious next step: growing the
corpus from 484K to 592K pairs at the fixed recipe \emph{regressed} both
benchmarks (bank crowding, \S\ref{sec:crowding}; partially recovered by the
soft-label and false-negative-masking terms), so corpus scale beyond
$\sim$500K pairs is not free under a full-corpus negative bank;
AuroLA's scaling curve, measured without one, suggests headroom in the
0.5M--1.4M range~\citep{aurola2026}. At fixed 2B base size the measured
levers are otherwise spent: connector capacity has a knee at 384
(\S\ref{sec:capacity}), caption rewriting and tower substitution reduce
accuracy (\S\ref{sec:negative}), and the fine-tune schedule is at its
optimum from both sides (\S\ref{sec:ftlength}). The headroom we can
identify: additional in-domain data, and the adapter rank axis, which had
not saturated at rank 384. The first lever has since been confirmed: after
this report's evaluations were frozen, a 1{,}600-step fine-tune on the
AudioCaps~2.0-expanded pool (86{,}394 pairs; train split verified disjoint
from our evaluation protocol) lifted AudioCaps \ratk{10} to 0.759
audio-to-text and 0.785 text-to-audio with Clotho and VGGSound within
noise, and is released as v0.2 of generation~2; all numbers reported here
remain the pinned v0.1 checkpoints.
Systems with native-audio backbones and in-backbone adaptation sit
$\sim$15 \ratk{1} higher on AudioCaps, which bounds what this family can
reach if the byte-identity constraint were relaxed, though OEA's
observation that its 7B backbone matches its 3B cautions against expecting
backbone scale alone to close it.

Three further limitations. \emph{Speech and music are not yet trained to
parity}: the instruction taxonomy includes speech content, language,
paralinguistics, and music tasks, but the corpus is sound-event-centric,
and MAEB's speech tasks are the models' weakest. \emph{English-only, with
input constraints}: captions are English; audio is 16\,kHz mono, processed
in 30\,s windows (longer audio chunked, up to 8 windows). \emph{Evaluation
breadth}: cross-modal results are on one benchmark (VGGSound-696) with a
single readout convention; MMEB-V2 preservation is mechanical (parameter
identity) rather than re-measured; and some baseline numbers (M2D-CLAP
direction labels, GLAP) are single-sourced, as flagged in the tables.
Finally, the training mix includes YouTube-sourced and research-licensed
corpora, so the preview checkpoints are released under CC-BY-NC-4.0; a
commercially clean corpus is a separate, ongoing track.

\section{Conclusion}
\label{sec:conclusion}

A frozen vision--language embedding base can be extended to audio without
touching a single base weight: first with a 16.4M connector trained in
five GPU-hours (\fea: AudioCaps \ratk{10} 0.741 at the frontier of the
frozen-both-towers class, strictly zero-shot Clotho transfer, emergent
audio--image retrieval at 29$\times$ chance), then with modality-gated
deep adapters that add in-layer capacity while keeping text, image, and
video outputs bit-for-bit identical to the released base (\feb: the
family's best text-to-audio results on every benchmark). Along the way we
isolated a training-protocol effect (+14.5 \ratk{10}) that we believe
generalizes to any frozen decoder-LM embedding backbone; a scaling caveat
for full-corpus negative banks past 500K rows; and controlled negative
results (caption rewriting, tower substitution, and connector widening
all \emph{reduce} accuracy) that map where the headroom of
frozen-backbone audio fusion actually lives, and that generation~2
converts into gains. Future work follows directly: speech and music data
to make the instruction taxonomy real, the unsaturated adapter-rank axis,
and the same recipe on the 8B
base tier, primarily for the quality of the shared text/image/video
space that audio binds into.

Both generations are released: connector and adapter checkpoints with
pinned revision tags at
\url{https://huggingface.co/EximiusLabs/fusion-embedding-1-2b-preview}
(\texttt{v0.1}/\texttt{v0.2}/\texttt{v0.3-preview}) and
\url{https://huggingface.co/EximiusLabs/fusion-embedding-2-2b-preview}
(\texttt{v0.1/v0.2-preview}), with Apache-2.0 training and evaluation code
at
\url{https://github.com/Eximius-Labs/fusion-embedding}. The generation-1
model is integrated in the \texttt{mteb} library, loadable by name through
\texttt{mteb.get\_model} with plain \texttt{transformers} remote code.

\section*{Reproducibility statement}
The training and evaluation pipeline is open source
(\url{https://github.com/Eximius-Labs/fusion-embedding}), and preview
checkpoints ($\sim$60\,MB connector-only; the frozen towers download from
their original repositories) are released with pinned version tags
corresponding to the checkpoints evaluated here. The generation-1 model is
integrated in the \texttt{mteb} library, so the MAEB evaluations of
\S\ref{sec:breadth} reproduce end-to-end from the public package. Architecture and
hyperparameters are specified in
\S\ref{sec:method}--\S\ref{sec:data} and Appendix~\ref{app:hyper};
evaluation protocols and their data sources (the AudioCaps test crawl, the
Clotho v2.1 Zenodo release, and the public VGGSound retrieval pairing) in
\S\ref{sec:protocols}. Every training run emits a result record containing
its configuration, corpus size, loss trace endpoints, the
\texttt{base\_drift} assertion, and an automatically scored AudioCaps
protocol result; checkpoints record their base precision so evaluations
cannot silently mix precisions. The full pipeline also runs end-to-end on
CPU-only stand-ins with over 125 unit and integration tests, including
the bitwise-invariance and gate tests of \S\ref{sec:invariance}, so the
training logic can be verified without GPUs.

\section*{Acknowledgments}
This work builds on outstanding open releases: Qwen3-VL-Embedding and
Qwen2.5-Omni (the frozen towers), ImageBind's emergent-alignment result,
Matryoshka representation learning, and the open audio--caption data
ecosystem (AudioCaps, WavCaps, Clotho, FSD50K, VGGSound, ESC-50, and
LAION-FreeSound).

\bibliography{refs}

\clearpage
\appendix

\section{Hyperparameters}
\label{app:hyper}

\begin{table}[h]
\centering
\footnotesize
\begin{tabular}{@{}l@{\hspace{0.9em}}l@{}}
\toprule
Connector $d_r$ / latents / blocks & 384 / 64 / 6 \\
Adapter rank / layers (Gen.~2) & 384 / 28 \\
Optimizer; schedule & AdamW; cos., 5\% warmup \\
Effective batch & 1024 (128 $\times$ 8) \\
Steps (G1 / G2 pretrain / FT) & 3200 / 3900 / 400 \\
Base precision & bf16 \\
Loss & Eq.~(1); $\lambda{=}0.05$ \\
Large-scale terms (500K+) & soft $\beta{=}0.3$, mask $\tau{=}0.98$ \\
Negative bank & full corpus, frozen text \\
Temperature init / clamp & $\log(1/0.07)$ / $\log 100$ \\
Retrieval rung & 1024-d \\
Audio window & 30\,s, 16\,kHz; 10\,s crops \\
\bottomrule
\end{tabular}
\caption{Training configuration for the released checkpoints
(\S\ref{sec:config}).}
\label{tab:hyper}
\end{table}

\section{Test-time hubness correction is benchmark-specific}
\label{app:qbnorm}

Querybank normalization (QB-Norm)~\citep{bogolin2022qbnorm} corrects
retrieval hubness at inference time by renormalizing gallery similarities
against a bank of query embeddings. We built the querybank from training
captions and used the dynamic inverted softmax variant, selecting the
temperature ($\beta = 20$) on the AudioCaps benchmark, where it adds
+1.75 text-to-audio \ratk{10} (0.7506 $\to$ 0.7681). All numbers in this
study are the generation-1 v0.3 checkpoint under the study's own
evaluation run, whose baselines differ slightly from the release-protocol
cells of Table~\ref{tab:audiocaps}. Applied \emph{once},
with no further tuning, to Clotho it adds +0.9 text-to-audio \ratk{1} and
+1.0 text-to-audio \ratk{10} (0.4624 $\to$ 0.4720); applied to a third
benchmark (MECAT) it \emph{hurts}: $-1.2$ \ratk{1} and $-1.7$ \ratk{10}.
We therefore ship it as an optional inference-time setting, off by default,
and no headline number in this paper includes it. The sign flip is a
useful caution: test-time corrections tuned on one benchmark do not
transfer as reliably as their single-benchmark gains suggest.

\section{Adapter gate probe}
\label{app:probe}

The controlled experiment behind generation~2 (matched 45K AudioCaps-only
arms, 800 steps, identical recipe, in-run protocol): control (no adapters)
\ratk{1} 0.202 / \ratk{5} 0.484 / \ratk{10} 0.631, text-to-audio \ratk{10}
0.684; adapters rank 128 (14.8M) 0.217 / 0.512 / 0.656; adapters rank 384
(44.2M) \textbf{0.229} / \textbf{0.531} / \textbf{0.665}, text-to-audio
\ratk{10} \textbf{0.708}, with a lower final training loss. The
acceptance gate was $\geq$+3 \ratk{10}; it passed at +3.4, every direction
improved, and the gain had not saturated at the largest rank trained. At
full scale the pretraining-stage gain reproduced at the same magnitude
(\ratk{10} 0.708 with adapters at 518K with the large-scale loss terms
vs.\ 0.674 for the 592K baseline without either, matched protocol; the
probe above is the controlled isolation of the adapters themselves).

\section{Full AudioCaps retrieval grids}
\label{app:grids}

\begin{table}[h]
\centering
\footnotesize\setlength{\tabcolsep}{4pt}
\begin{tabular}{lcccc}
\toprule
 & v0.1 & v0.2 & v0.3 & Gen.~2 \\
\midrule
A$\to$T \ratk{1} & 0.216 & 0.279 & 0.332 & 0.302 \\
A$\to$T \ratk{5} & 0.479 & 0.590 & 0.602 & --- \\
A$\to$T \ratk{10} & 0.626 & 0.717 & 0.741 & 0.743 \\
A$\to$T mAP@10 & 0.159 & 0.208 & 0.237 & --- \\
T$\to$A \ratk{1} & 0.226 & 0.268 & 0.280 & 0.292 \\
T$\to$A \ratk{5} & 0.530 & 0.589 & 0.602 & --- \\
T$\to$A \ratk{10} & 0.680 & 0.736 & 0.746 & 0.775 \\
T$\to$A mAP@10 & 0.354 & 0.404 & 0.418 & --- \\
\bottomrule
\end{tabular}
\caption{Released-checkpoint AudioCaps grids under the release protocol
(\S\ref{sec:protocols}). Generation-1 cells that were not tabulated at
release time were recovered from the archived release-time evaluation
artifacts (each artifact reproduces every previously published cell
exactly); no checkpoint was re-scored. Remaining dashes: not recorded.}
\label{tab:grids}
\end{table}

\section{Negative-result tables}
\label{app:negtables}

\begin{table}[H]
\centering
\footnotesize\setlength{\tabcolsep}{3.5pt}
\begin{tabular}{lcc}
\toprule
Arm & AC \ratk{10} & Clotho \ratk{10} \\
\midrule
Rewritten (518K + loss terms) & 0.649 & 0.438 \\
Raw, same scale (592K, no terms) & 0.674 & 0.400 \\
Raw, 484K (v0.2, context) & 0.717 & 0.448 \\
\bottomrule
\end{tabular}
\caption{Recaptioning arms (\S\ref{sec:recaption}), audio-to-text. The
592K row is the same-scale baseline; the loss terms favor the rewritten
arm by +2.6 (\S\ref{sec:crowding}), so $-2.5$ under-states the rewrite's
cost. Clotho baseline cells are from the run logs.}
\label{tab:recaparms}
\end{table}

\begin{table}[H]
\centering
\footnotesize\setlength{\tabcolsep}{3.5pt}
\begin{tabular}{lcc}
\toprule
Tower & A$\to$T \ratk{10} & T$\to$A \ratk{10} \\
\midrule
Qwen2.5-Omni ($\sim$640M) & \textbf{0.631} & \textbf{0.684} \\
Dasheng-base (86M) & 0.469 & 0.538 \\
\bottomrule
\end{tabular}
\caption{Tower-swap A/B (\S\ref{sec:towerswap}): 45K AudioCaps-only
corpus, 800 steps, identical recipe, each arm on frames from its own
tower. Text-to-audio cells are from the run logs. The towers are not
parameter-matched; the experiment establishes non-transfer of
shallow-projection rankings, not encoder superiority per parameter.}
\label{tab:towerswap}
\end{table}

\section{Invariance verification protocol}
\label{app:invproto}

The checks behind \S\ref{sec:invariance}, by artifact:
\texttt{fusion\_embedding/adapters.py} implements the gate (a depth-counted
context manager) and the layer hooks, which return the frozen layer's
output \emph{before} any adapter arithmetic when the gate is closed.
\texttt{tests/test\_adapters.py} asserts: (i) bitwise text invariance:
identical activations through the base and through the adapter-attached
model with the gate closed, and exact identity of a zero-initialized
adapter stack under an open gate; (ii) audio-gated forwards \emph{change}
while a text forward in the same process remains bit-identical; (iii)
gradient isolation: only adapter and connector parameters receive
gradients, and the frozen towers reject them; (iv) checkpoint resume
round-trips adapters and refuses adapter-presence mismatches; (v)
warm-start semantics from a connector-only checkpoint; (vi) a gate that
fails to span the re-run forwards of gradient checkpointing raises
loudly (\texttt{CheckpointError}) rather than silently degrading the
guarantee. The training loop asserts \texttt{base\_drift} $=0$ (maximum
absolute parameter change) at the end of every run, and the release smoke
test loads the packaged checkpoint from the public repository, repeats the
text output-equality check, and exercises the audio, text, and image paths
end-to-end through the public loading code.

\section{Input formats, verbatim}
\label{app:templates}

The native chat-template format (the base's official protocol;
\S\ref{sec:native}), with the task instruction in the system turn and
pooling at the final position:

\begin{quote}
\footnotesize
\begin{verbatim}
<|im_start|>system
{instruction}<|im_end|>
<|im_start|>user
{caption}<|im_end|>
<|im_start|>assistant
\end{verbatim}
\end{quote}

\noindent
The bare format used by our early runs (off-manifold for the frozen
base; $-14.5$ \ratk{10} at matched configuration):

\begin{quote}
\footnotesize
\begin{verbatim}
{instruction} {caption}<|im_end|>
\end{verbatim}
\end{quote}

\noindent
An example instruction (the sound-retrieval task): ``Retrieve audio by
sound description.'' Sequences are truncated at 254 tokens; the embedding
is pooled at the last position in both formats.

\clearpage
\section{MAEB per-task scores}
\label{app:maeb}

\begin{table}[h]
\centering
\footnotesize
\begin{tabular}{lcc}
\toprule
Task & Main score & Rank \\
\midrule
BeijingOpera (cls.) & 0.928 & \#6 \\
ClothoT2ARetrieval & 0.278 & \#12 \\
GTZANAudioReranking & 0.707 & \#16 \\
GTZANGenre (cls.) & 0.639 & \#29 \\
MACST2ARetrieval & 0.132 & \#14 \\
RavdessZeroshot & 0.322 & \#4 \\
SpeechCommandsZeroshot v0.02 & 0.190 & \#11 \\
UrbanSound8KT2ARetrieval & 0.009 & \#3 \\
VehicleSoundClustering & 0.035 & \#17 \\
\bottomrule
\end{tabular}
\caption{MAEB sound-event tier, \fea{} v0.2-preview with
\texttt{mteb==2.18.0}: each task's \texttt{main\_score} as defined by the
benchmark (metrics differ per task type), with public-leaderboard ranks
as of 2026-07-09. Absolute scores on UrbanSound8KT2ARetrieval are low
across the leaderboard (best published 0.010); the rank, not the
magnitude, is informative there. FSD2019Kaggle withheld for the
containment issue described in \S\ref{sec:breadth}.}
\label{tab:maeb}
\end{table}

\end{document}